# Beyond Benchmarks: Exposing the Hidden Crisis in Bangla Hate Speech Detection

**Faria Afrin Tisha[1], Fariya Tabassum[1], Hafsa Binte Kibria[1], Md. Nahiduzzaman[1] and Mominul Ahsan[2*]**

[1]Department of Electrical and Computer Engineering, Rajshahi University of Engineering and Technology, Rajshahi, 6204, Bangladesh.
email: 1810017@student.ruet.ac.bd (F.A.T); tabassum.fariya27@gmail.com (F.T); hafsabintekibria@ece.ruet.ac.bd (H.B.K); nahiduzzaman@ece.ruet.ac.bd (M.N.).
[2]Department of Computer Science, University of York, Deramore Lane, Heslington, York YO10 5GH, UK; email: mominul.ahsan2@gmail.com (M.A.).

*Corresponding author: mominul.ahsan2@gmail.com (M.A.)

**Abstract**

The spread of hate speech (HS) across different social media platforms (SMPs) poses a major concern for ensuring online safety and ethical moderation. Automatic detection of HS remains a challenging task, especially in under-resourced languages like Bangla, due to cultural context, implicit expressions, and informal linguistic patterns. This study aimed to expose the underlying crisis of existing Bangla HS detection systems by diagnosing how and why benchmark-trained models fail to identify implicit, context-dependent HS. Six deep learning architectures (FastText + CNN, FastText + LSTM, FastText + BiLSTM, BanglaBERT, BanglaBERT + CNN, and BanglaBERT + BiLSTM) were trained on benchmark datasets (≈75,000 posts) and a merged multi-source dataset (≈120,000 posts), then externally validated on a newly annotated real-world dataset (≈200 posts) collected from Facebook, Twitter, and YouTube, labeled as HS and non-HS, where HS was further categorized as explicit and implicit. BanglaBERT achieved an F1-score of 91.4% on benchmark datasets but declined to 75.3% on the external set and 63.4% for implicit HS involving sarcasm and emojis. The accuracy of FastText + CNN dropped from 78.0% to 51.2% under similar conditions. Emoji-aware preprocessing improved implicit HS detection by up to 12%, whereas emoji removal caused a notable decline in performance (F1: 0.75 → 0.63). Frequent misclassifications in politically charged or satirical comments revealed over-policing risks. This study not only exposes the generalization crisis due to implicit, culturally embedded, and emoji-laden expressions but also underscores the need for developing adaptive, emoji-aware, and culturally grounded frameworks that ensure ethical moderation while preserving freedom of expression. Findings of this study provide actionable insights for researchers, SMPs, and policymakers to design more context-sensitive HS detection systems for low-resource languages.

## 1 Introduction

Over the past decade, SMPs such as Facebook, YouTube, Twitter, and TikTok have changed the game of public communication by providing unprecedented room for freedom of expression. However, this freedom has also allowed cyberbullying (Ray et al., 2024) in the form of HS. HS refers to a direct or indirect attack on individuals based on protected qualities such as race, ethnicity, nationality, disabilities, religious affiliation, caste, gender, and serious disease, which is severe enough to cause harm by dehumanizing words or negative stereotypes (*Hateful Conduct | Transparency Centre*, n.d.) People or groups are targeted based on these qualities. The prevalence of social media services in Bangladesh has resulted in a disturbing rise in HS, which poses a serious threat to social harmony, personal security, and internet well-being (Atikuzzaman & Akter, 2023), While languages with high resources, such as English, and Arabic, have experienced tremendous success in detecting HS through machine learning (ML) and deep learning (DL), languages such as Bangla remain critically underexplored (Albladi et al., 2025). This task is challenging for several reasons. First, there is a lack of adequate annotated corpora, benchmark datasets, and linguistic resources necessary for large-scale model training.

Second, the character of HS in online discussions is culturally subtle, highly contextual, and frequently implicit. Users frequently express hate using sarcasm, memes, coded language, or emojis communication types that evade straightforward lexical or keyword-based detection approaches (Ghenai et al., 2025). Accordingly, models that perform so effectively on benchmark datasets will collapse or be neglected to spot actual HSs in casual and emotionally intense online settings. In recent years, several Bangla HS detection models have emerged, ranging from traditional ML and DL methods to transformer-based pretrained language models such as BanglaBERT (Bhattacharjee et al., 2022). These models are primarily evaluated on benchmarked datasets such as BD-SHS (Romim et al., 2022), NNTI (Romim, Ahmed, Islam, et al., 2021), and SentNoB‘ (Islam et al., 2021) collected by the researchers. However, how well these models generalize to real, diverse, and conversational Bangla communication, particularly on dynamic SMPs, has been less investigated.

Unlike prior comparative works, the present study does not solely aim to report performance accuracy. Instead, it seeks to uncover the deeper reasons behind the persistent misclassification of implicit HS, that is, hate expressed indirectly through contextual, sarcastic, or emoji-based cues. To achieve this, a newly annotated external validation dataset was constructed from real-world Bangla SMP content, labeled as hate and non-hate, with hate further categorized as explicit and implicit. This external dataset served not as training data but as an independent diagnostic tool to evaluate and interpret how benchmark-trained models behave on unseen, context-dependent expressions.

This work aimed to bridge these gaps with a systematic comparative analysis of existing Bangla HS detection models. We investigated whether the best-performing models on benchmark sets could generalize towards implicitly hateful real-world SMP content. This work is inspired by the following research questions:

1. Do high-performing Bangla HS detection models on benchmark datasets generalize to implicitly hateful real-world content?
2. How do emojis influence the understanding and classification of implicit HSs, and are models capable of effectively capturing the emotional or sarcastic, ambiguous cues they convey?
3. How does over-policing Bangla HS risk freedom of speech?

To answer these questions, a novel comprehensive comparative study of six deep learning models was conducted using FastText and BERT-based embeddings. These models were evaluated on three categories of datasets: (i) benchmark datasets commonly referenced in the literature, (ii) a merged dataset consisting of several sources for enhanced representation, and (iii) an external, manually annotated Bangladeshi social media post dataset labeled for both explicit and implicit HS. Our findings showed a significant decline in performance when models trained on typical datasets were applied to real, informal, dynamic, and culturally complex content. Explicit HSs are properly identified at an adequate rate, whereas models badly fail in implicit cases such as sarcasm or culture-coded messages. This study included different preprocessing techniques such as emoji translation and summarization that can affect the detection accuracy, especially for indirect HS. The significance of emojis is especially evident, whereas their emotional connotations or ironic intentions can mislead or augment model understanding based on how they are processed during preprocessing. The novelty of this work lies in two major aspects:

- **Context-Aware and Emotionally Sensitive Approach:** Unlike the conventional methods that only use benchmark datasets or keyword detection, the generalization of the Bangla HS detection models in real-life informal, culturally vibrant, and evolving social media content is explored. We contribute by using implicit HS with emojis and colloquial Bangla language, which is a novelty in bridging the gap of culturally and emotionally aware methods for detecting HS.
- **Comprehensive Comparison:** A detailed comparison was conducted using six state-of-the-art deep learning models, which included models with FastText as well as BERT embedding. This study is more than a new dataset but is about how pretrained models (which are normally evaluated on stringent, organized, rigorously annotated benchmark datasets) perform on real-world noisy user-generated SMP content. This paper, thus, provides an understanding of the limitations of the existing Bangla HS detection systems and presents clear directions to develop culturally aware and contextually sensitive detection systems, especially for low-resource languages such as Bangla. These findings highlight the limitations of current Bangla HS detection systems and the necessity of context-dependent, culture-sensitive modeling approaches. This paper outlines those limitations, presents complete experimental results, and outlines directions for future work in improving the robustness and portability of HS detection for low-resource languages such as Bangla.

While the above research questions define the investigative scope, the study’s purpose extends beyond quantitative evaluation to a broader diagnostic inquiry. It aims to reveal the underlying weaknesses of benchmark-based Bangla HS detection systems and to provide evidence-driven insights into how cultural context, implicit intent, and emoji semantics shape misclassification.

## 1.1 Research Objectives

Despite achieving high accuracy on benchmark datasets, existing Bangla HS detection models often fail to generalize to real-world social media content, particularly when hate is expressed implicitly through sarcasm, emojis, or culturally coded language (Fortuna et al., 2021). This limitation raises critical questions about the reliability and ethical application of current models in real online environments. To address these concerns, this study adopts a diagnostic approach rather than proposing a new model to uncover how and why benchmark-trained Bangla HS detection systems misclassify implicit HS. The overarching goal of this research is to expose the generalization crisis in automatic Bangla HS detection by identifying both the extent and the causes of model failure in real-world contexts. To achieve this, the study is guided by the following specific objectives:

1. To evaluate the performance of six state-of-the-art Bangla HS detection models, trained on benchmark and merged datasets, and when externally validated on real-world SMP content labeled as explicit and implicit HS.
2. To analyze linguistic, emotional, and contextual factors (e.g., sarcasm, emoji usage, and cultural references) that contribute to misclassification in implicit HS.
3. To identify the theoretical and practical implications of these findings, emphasizing the need for adaptive, emoji-aware, and culturally grounded frameworks for ethical, context-sensitive moderation in low-resource languages.

These objectives collectively aim to reveal not only how well models perform but also why they fail, thereby contributing diagnostic insights toward the development of more robust and socially responsible hate speech detection systems.

# 2 Related Works

Hate speech (HS) detection has emerged as a critical task in the intersection of natural language processing (NLP), social computing, and information management. Over recent years, there has been a shift in methodologies, from traditional machine learning (ML) approaches to deep neural networks and transformer-based models, which have made notable advances. However, despite these breakthroughs, HS detection systems face persistent challenges when applied to the noisy, informal, and culturally specific content encountered on social media platforms (SMPs), particularly in low-resource languages such as **Bangla**. In this section, we summarize the literature surrounding (i) HS detection in high-resource versus low-resource languages, (ii) the evolution of transformer and large language models (LLMs), (iii) cross-dataset generalization and external validation strategies, and (iv) the growing emphasis on implicit HS, such as sarcasm and emoji-laden content.

## 2.1 HS Detection in High-Resource vs Low-Resource Languages

Early efforts in HS detection relied heavily on handcrafted feature engineering and traditional classifiers such as SVM and LR, often coupled with TF-IDF or n-gram features (Fortuna & Nunes, 2019; Zhang et al., 2018). These methods were most effective when large, well-annotated corpora were available. However, as highlighted by recent systematic reviews, the performance of these models is strongly constrained by dataset quality and annotation practices, particularly for low-resource languages like Bangla. Low-resource languages face unique challenges, such as a lack of annotated data, high dialectical variation, and informal linguistic expressions often found in user-generated content on SMPs. This leads to significant issues with the generalization of models trained on well-curated, formal datasets to the noisy, dynamic environments of social media. In particular, research from ScienceDirect and other outlets has pointed out the dataset bias in current approaches, where models trained on specific datasets may not perform well on others due to variations in text structure and annotation guidelines . In low-resource settings, there is an urgent need for cross-dataset validation to assess a model's ability to generalize to real-world content.

Subsequent work leverages DL methods, including Convolutional Neural Networks (CNNs), RNNs, and transformer models such as BERT, RoBERTa, and XL-Net, that are pre-trained on large corpora and then fine-tuned for HS classification (Badjatiya et al., 2017; Gambäck & Sikdar, 2017; Liu et al., 2019; Mozafari et al., 2019; Yang et al., 2019; Zhang & Luo, 2019). Rule-based filtering and DL methods like AraBERT and BiLSTM have been applied in the case of Arabic to handle dialectical heterogeneity and code-switching problems with state-of-the-art performance on tasks related to offensive and abusive language (Koshiry et al., 2023; Mubarak et al., 2021). For low-resource South Asian languages such as Hindi, Urdu, and Tamil, researchers have highlighted transfer learning from multilingual models (e.g., mBERT, XLM-R) and resource-augmenting practices such as cross-lingual pretraining or translation-based synthetic data generation (De et al., 2021; Mnassri et al., 2024). They share similar challenges as Bangla does, including orthographic variation, social media texts in informal environments, and sparse labeled corpora. Projects such as the MUHSIN corpus for Hindi and abusive language detection in Tamil Twitter data demonstrate the emerging but limited work here (De et al., 2021).

## 2.2 Bangla HS: Datasets and Benchmarking Progress

Recent work on Bangla hate speech detection has progressed significantly, largely due to the creation of benchmark datasets such as HS-BAN (Islam et al., 2021; Romim, Ahmed, Islam, et al., 2021), a massive collection of binary-class HSs with more than 50,000 Bangla comments labeled as hateful or not hateful, along with annotations for the usage of slang and category-specific text such as religion, politics, and sports. Similarly, the BD-SHS(Romim et al., 2022) dataset includes 30,000 user comments posted on Facebook and YouTube, which are labeled in seven thematic domains. Both datasets served as a reference in various subsequent studies. Some of the other notable contributions include the 250 million large-scale Bangla article corpus for Bangla word embeddings, which utilized the BengFastText model to train (Karim et al., 2020), and the BHSSP dataset (Asad et al., 2025) comprising over 110,000 insulting words collected from Bangladeshi news portals and websites. Another dataset, SentNoB‘ (Islam et al., 2021), is a 15,000 human-annotated dataset of casual Bangla SMP messages that enables sentiment analysis in noisy, dialect-rich, and grammatically-unstable text environments. These have contributed to the growth of Bangla NLP, despite the language being a low-resource language, particularly those based on DL architectures.

For Bangla, previous studies employed ML classifiers such as SVM and Naïve Bayes with TF-IDF and bag-of-words features (Ishmam & Sharmin, 2019). While they were effective, they lacked contextual data to comprehend informal Bangla or ambiguous hate statements. Later studies began investigating deep learning architectures such as CNN, Bidirectional Long Short-Term Memory (Bi-LSTM) networks, and hybrid models that combined word embeddings such as Word2Vec and FastText (Banik & Rahman, 2019). For instance, when training embedding models with in-domain slang, FastText and Bi-LSTM performed well on HS-BAN's informal text with an F1 score of 86.78%. Recent efforts have focused on transformer-based models, with BanglaBERT taking state-of-the-art by pre-training on 27.5 GB of Bangla text (Bangla2B+) (Bhattacharjee et al., 2022). The contextualized embeddings of BanglaBERT have been employed for HS detection, which outperformed earlier methods in both benchmark and real-world datasets. Other notable models are BanglaHateBERT, which was trained solely on abusive text, and DeepHateExplainer, which employed explainable AI (XAI) layers to improve interpretability, but at the cost of generalizability (Karim et al., 2021). Hybrid approaches that combine BanglaBERT with CNN or BiLSTM layers have also been proposed to identify contextual and sequential patterns in HS detection (Das et al., 2021). Some works have also proposed fuzzy SVMs and morphological analyzers to handle lexical uncertainty and class imbalance—a few of the most significant issues in Bangla HS detection (Ghosal et al., 2023). However, all of these models still rely most strongly on benchmark sets and fail to generalize across implicit, emoji-based, and culturally encoded hate found on Bangladeshi social media. Our work builds on this by providing a comparative analysis of six architectures (FastText and BanglaBERT variants) across benchmark, merged, and external sets, specifically focusing on implicit hate, emoji comprehension, and issues of over-policing.

## 2.3 Transformer/LLM Era: Gains and Limitations

The introduction of transformer-based models like BERT marked a significant leap in HS detection, enabling models to capture deeper semantic and contextual cues in text. Monolingual models such as BanglaBERT have shown superior performance compared to multilingual alternatives, suggesting that language-specific pretraining can dramatically enhance downstream performance on tasks such as Bangla HS detection (Mahmud et al., 2023). However, the rise of large language models (LLMs), such as GPT-3 and GPT-4, has complicated the landscape. Although LLMs, particularly those utilizing prompt-based learning, have achieved impressive performance on standard benchmarks, evidence suggests that they struggle with code-mixing, dialectal variations, and culturally encoded implicit hate when the pretraining corpus under-represents certain target languages (Dmonte et al., 2024). For instance, GPT-3 and GPT-4 models, trained on large, diverse corpora, often fail to handle the subtleties of Bangla HS, especially when the content includes sarcasm, emoji-laden speech, or implicit hate. This limitation has been particularly evident in cases where Bangla HS is conveyed indirectly via sarcasm, irony, or humor, which are common in Bangladeshi social media discourse.

## 2.4 Cross-Dataset Generalization and Diagnostic Analyses

Recent literature has emphasized the importance of cross-dataset evaluation to assess how models trained on one dataset perform on others. Studies such as (Romim, Ahmed, Talukder, et al., 2021) and BanglaHateBERT (Jahan et al., 2022) have illustrated that models trained on benchmark corpora often show a significant drop in performance when evaluated on real-world or externally collected datasets. This is largely due to the mismatch in text characteristics—such as informality, dialectical variations, and cultural references, between benchmark datasets and the dynamic, noisy text found in real social media platforms. Models often suffer from overfitting to dataset-specific cues (e.g., common keywords or platform-specific slang), which hinders their ability to generalize to new, unseen data (Fortuna et al., 2021). This underscores the need for external validation using datasets that capture a broader variety of real-world content and culturally specific expressions. In this research, models like BanglaBERT perform well on benchmark datasets but show a sharp performance decline when faced with implicit hate speech containing sarcasm or emojis.

## 2.5 Implicit Hate, Sarcasm, and Emojis: Under-Studied but Decisive

The detection of implicit hate speech—expressed through sarcasm, irony, memes, or culturally coded language—remains a significant challenge. While explicit hate speech is typically more straightforward to identify, implicit HS often relies on contextual, cultural, and emotional cues that are difficult for models to detect without a deeper understanding of the language (ElSherief et al., 2021). Recent studies have demonstrated that emojis play a critical role in shaping the meaning of a text, and their omission can significantly degrade a model's performance. Our work emphasizes the importance of emoji-aware preprocessing for improving the detection of implicit HS. The research finds that emoji removal leads to a substantial performance drop, particularly for implicit HS, while translating emojis into sentiment-rich words or phrases enhances model performance. The ability of BanglaBERT to process emoji-laden content, in particular, has proven superior to LLMs like GPT-3 and GPT-4, which are not specifically trained to interpret the cultural and emotional subtleties conveyed through emojis.

## 2.6 Ethics, Moderation, and Over-Policing Concerns in Information Science

Automated content moderation systems for HS detection raise several ethical concerns, particularly the risk of over-policing legitimate speech, especially in politically sensitive or satirical contexts. Studies in information science argue that ethically grounded models are crucial to avoid suppressing free speech or democratic participation, particularly in low-resource languages like Bangla, where language nuances can be easily misinterpreted. BanglaBERT performed well on explicit HS but faced challenges when it came to implicit HS, such as political satire or humor. This reveals the potential for over-policing, where content that should be classified as non-hateful is flagged as harmful due to misinterpretation of sarcasm or coded language. Thus, it is essential that models incorporate contextual grounding and cultural awareness to avoid such ethical pitfalls.

## 2.7 Gap Analysis

Although there has been substantial progress in Bangla HS detection, key gaps remain: (1) benchmark bias, where models trained on curated corpora perform poorly in real-world settings; (2) implicit hate speech, which is underrepresented in current datasets; and (3) lack of diagnostic external validation to assess why models fail in real applications. The findings from your research align with these gaps, emphasizing the need for emoji-aware preprocessing, contextual models, and external validation sets to improve the robustness and fairness of HS detection systems. The present work approach training models on merged datasets and evaluating them on real-world, emoji-rich, and implicitly hateful content addresses these gaps, offering insights into the limitations of current models and providing actionable recommendations for future improvements.

**Table** 1: Summary of Related works on Bangla HS detection.

| **Reference** | **Core Model(s)** | **Highest Reported Score** | **Dataset / Size (approx.)** | **Label Categories** | **Platform / Source** | **Contribution** |
|---|---|---|---|---|---|---|
| (Ishmam & Sharmin, 2019) | GRU, SVC, RF, NB | 0.70 (A) | ≈ 5 k comments | Hate (19 %), Inciteful (11 %), Religious hatred (16 %), Political (23 %) | Facebook | First Bangla HS corpus with multi-topic labels; establishes early baselines. |
| (Karim et al., 2021) | LR, SVM, CNN, Bi-LSTM, Conv-LSTM, BERT | 0.91 (F1, BERT) | ≈ 8 k posts | Personal, Political, Religious, Geo-political | Facebook | Compared classical vs deep vs transformer models; highlights BERT gains. |
| (Sazzed, 2021) | SVM, LR, RF, Bi-LSTM | 0.83 (F1) | ≈ 3 k comments | Abusive / Non-abusive | YouTube | Shows deep learning outperforms feature-based ML in small, noisy data. |
| (Romim et al., 2022)– *BD-SHS* | LSTM, Bi-LSTM | 0.88 (A) | ≈ 30 k comments | Hate (33 %) / Non-hate (67 %) across 7 domains | Facebook, YouTube | Most widely used Bangla HS benchmark; diverse thematic coverage. |
| (Islam et al., 2021) | MNB, MLP, SVM, RF, SGD, k-NN | 0.88 (A) | ≈ 9.8 k posts | Abusive / Non-abusive | Facebook, YouTube | Multi-algorithm comparison; verifies classical ML ceiling on Bangla HS. |
| (Aurpa et al., 2021) | BERT, ELECTRA, Transformer-DNN | 0.85 (A, BERT) | ≈ 44 k posts | Sexual (20 %), Troll (24 %), Religious (17 %), Threat (4 %), Not-bully (35 %) | Facebook | Introduced domain-rich Bangla cyberbullying dataset; transformer baseline. |
| (Bhattacharjee et al., 2022)– *BanglaBERT* | Transformer (Pretrained BanglaBERT) | 0.92 (F1 on BD-SHS) | Bangla2B+ (≈ 2 B tokens pretraining) | (pretraining corpus) | Large web corpora | Monolingual Bangla pretraining yields +5 F1 vs mBERT; state-of-the-art baseline. |
| (Antypas & Camacho-Collados, 2023)– Cross-Dataset Study | BERT, RoBERTa, FastText | 0.86 (F1 in-domain), 0.63 (cross-domain) | ≈ 200 k (multilingual) | Hate / Offensive / Neutral | Twitter, Reddit | Exposes dataset bias and generalization loss → motivates external validation. |
| (Guo et al., 2024)– LLMs for HS | GPT-4, mT5, LLaMA-2 | 0.89 (F1 avg high-resource), ≤ 0.70 | ≈ 120 k (multi-source) | Hate vs Non-hate | Multiple benchmarks (HateXplain, | LLMs show strong few-shot ability but weak implicit HS handling in Bangla. |

| | | (low-re- source) | | | OLID, BD-SHS) | |
|---|---|---|---|---|---|---|
| (Ocampo et al., 2023)– Implicit HS Generation | Autoregressive LM + BERT Classifier | ≈ 20 pp drop on implicit HS vs explicit | ≈ 5 k generated examples | Explicit / Implicit HS | Twitter, Reddit | Demonstrates sarcasm and coded hate cause severe misclassification errors. |

## 3 Comparison to State-of-the-Art Baselines

In this study, we aimed to evaluate the generalization ability of existing state-of-the-art (SOTA) models for Bangla hate speech (HS) detection, specifically by investigating whether these models, which perform well on benchmark datasets, can generalize effectively to real-world social media posts. While large language models (LLMs) like GPT-2, GPT-3, and LLaMA have shown impressive results in certain NLP tasks, we did not choose to fine-tune or evaluate these models for several key reasons:

- **Model Task Specialization:** LLMs like GPT-2 and GPT-3 are general-purpose language models designed primarily for text generation, rather than task-specific text classification. While these models excel at generating coherent and contextually relevant text, they are not optimized for classification tasks like hate speech detection. Models such as BanglaBERT, on the other hand, are pre-trained on tasks like text classification and are inherently more suited for tasks where the goal is to predict a specific label (e.g., hate speech or not). In contrast, LLMs require significant modifications and prompt engineering to adapt them for classification tasks .
- **Performance on Implicit Hate Speech:** One of the primary motivations for this study was to evaluate how well benchmark-trained models generalize when exposed to implicit hate speech, such as sarcasm, cultural references, and emoji-laden content, which are prevalent in Bangla social media posts. While LLMs like GPT-3 are powerful in zero-shot or few-shot learning scenarios, they struggle significantly with context-dependent expressions, especially in low-resource languages like Bangla. Recent studies have demonstrated that GPT-3 and GPT-4 perform well in high-resource languages but fail to capture implicit hate speech, particularly in low-resource languages like Bangla, where sarcasm, irony, and emoji-driven sentiment play a crucial role in conveying hate speech . In contrast, BanglaBERT, fine-tuned on Bangla text, has shown superior performance in capturing contextual understanding, especially when it comes to implicit hate speech
- **Generalization Ability in Real-World Contexts:** Our main research goal was to evaluate how well benchmark-trained models generalize to real-world, informal Bangla content, which often includes implicit hate speech, sarcasm, and emoji use. While LLMs like GPT-3 have shown impressive capabilities in few-shot learning, they require substantial prompt engineering and fine-tuning to adapt to specific tasks like hate speech detection. In contrast, BanglaBERT has been fine-tuned specifically for classification tasks in the Bangla language and is more suitable for detecting subtle forms of hate speech that often occur in informal, noisy online environments. Fine-tuning GPT-3 or GPT-4 for this specific task would have required significant adaptation and might not have outperformed BanglaBERT due to these models' inherent limitations in implicit hate speech detection .
- **Practical and Computational Constraints:** The decision to not fine-tune LLMs like GPT-3 or LLaMA was also based on practical considerations. Fine-tuning LLMs for hate speech detection, especially for Bangla, requires extensive computational resources, large-scale pretraining, and a significant amount of labeled data. In contrast, BanglaBERT—which is specifically designed for tasks like hate speech detection has shown strong performance on Bangla-specific tasks with far less computational cost and data. Additionally, GPT-3 and LLaMA are large and require more fine-tuning data to perform adequately for low-resource languages like Bangla. This makes BanglaBERT a more efficient and effective solution, especially for real-world deployment, where computational efficiency and generalization to diverse content are paramount .

## *3.1 General Performance Results of LLMs*

Recent studies, such as the work by (Guo et al., 2024), have shown that while GPT-4 achieves strong performance in high-resource languages, its ability to detect implicit hate speech in low-resource languages like Bangla is considerably limited. These models generally achieve lower F1-scores (around 0.70) for implicit hate speech tasks, especially when the text involves sarcasm, irony, or emoji-driven sentiment. Our decision to use BanglaBERT was driven by its higher accuracy and better contextual understanding for Bangla-specific implicit hate speech. In conclusion, while LLMs like GPT-3 and LLaMA show promise for many NLP tasks, their application to Bangla hate speech detection, particularly for implicit speech, sarcasm, and emoji-laden content, is less effective compared to BanglaBERT. This is because BanglaBERT is optimized for text classification tasks and is pre-trained on Bangla text, making it more task-specific and computationally efficient for detecting both explicit and implicit hate speech in real-world, noisy data.

# 4 Methodology

This study conducted a comprehensive comparative analysis of six deep learning architectures for Bangla hate speech (HS) detection—three based on FastText embeddings (FastText + CNN, FastText + LSTM, FastText + BiLSTM) and three based on transformer-based embeddings (BanglaBERT, BanglaBERT + CNN, BanglaBERT + BiLSTM). The models were selected to represent both traditional and contextual embedding paradigms, enabling a fair assessment of their generalization capability on benchmark, merged, and real-world datasets. Model categories and training setup are discussed below:

- FastText-based models rely on static word representations enriched by character-level n-grams, which are sensitive to data cleanliness and lexical normalization. Hence, these models required extensive text preprocessing to optimize embedding consistency and reduce noise in informal Bangla text.
- BanglaBERT-based models, on the other hand, employ contextualized embeddings generated by the model's SentencePiece tokenizer, pre-trained on large-scale Bangla corpora. Consequently, any aggressive preprocessing (e.g., stop-word removal, stemming, or lemmatization) would disrupt the pretrained tokenization schema. Therefore, for all BanglaBERT-based architectures, preprocessing was deliberately kept minimal to preserve contextual and syntactic information.

All models were trained on the same benchmark and merged dataset for fair comparison and then evaluated on (i) benchmark corpora, (ii) the merged dataset itself, and (iii) an external, manually annotated real-world dataset containing both explicit and implicit HS.

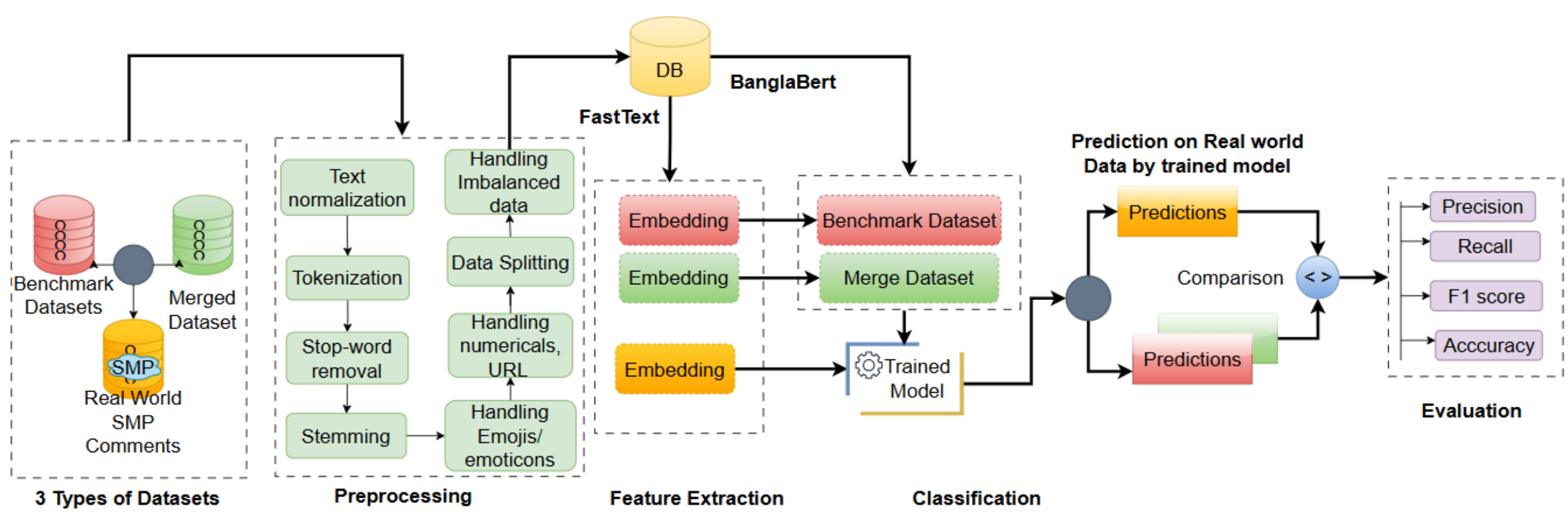

**Fig. 1** System pipeline for Bangla HS Detection

## 4.1 Datasets

This study utilized three categories of datasets to ensure a comprehensive and diagnostic evaluation of Bangla HS detection models. The first category includes benchmark datasets, such as BD-SHS, NNTI, and SentNoB, which represent the standardized corpora commonly used in prior Bangla HS research. These datasets were used for model training and initial testing to replicate state-of-the-art (SOTA) performance under conventional conditions. The second category, referred to as the merged dataset, was created by combining multiple benchmark datasets into a single, larger corpus to enhance linguistic diversity, topic balance, and vocabulary coverage. This merged set was also used for training and cross-validation to observe whether broader data representation improves model robustness. The third and most distinctive category is the external validation dataset, which was newly constructed from real-world Bangla SMP posts. Unlike the benchmark and merged datasets, this collection was not used for model training. Instead, it served as an independent evaluation set to examine how and why benchmark-trained models fail to generalize to implicit, context-dependent, and emoji-rich hate speech found in everyday online communication. Each dataset category is discussed in detail in the following subsections, outlining data composition, preprocessing steps, and its specific role in the comparative evaluation process.

### 4.1.1 Benchmark Datasets

A range of publicly available Bangla HS and sentiment corpora were utilized as baselines for training and testing. An overview of all major benchmark datasets given inside → Table 1. These datasets are employed to evaluate Bangla HS and sentiment analysis. Among them, three major benchmark datasets—NNTI (Romim, Ahmed, Talukder, et al., 2021), DeepHateExplainer (Karim et al., 2021), and BD-SHS (Romim et al., 2022), were employed to conduct the experiments. Contextualized examples and structures of the three selected datasets are illustrated in (→ Tables 2., 3., and 4.), respectively. These resources store different sources such as Facebook, YouTube, Twitter, and news websites on the internet, covering a number of topics such as religion, politics, sports, and individuals.

**Table 2**: Existing Bangla HS and Sentiment Corpus

| Dataset | Size (Posts) | Platforms | Categories | Tool Used |
|---|---|---|---|---|
| Bengali Language Corpus for HS [23] | 5,000 | Facebook | HS, Inciteful, Religious, Religious Hatred, Political Comment, Communal Hatred | Facebook Graph API |
| Corpus for HS (Large and Diverse) [6] | 30,000 | YouTube and Facebook | Sports, Entertainment, Crime, Religion, Politics, Celebrity, Meme, TikTok | Facepager [a] |
| NNTI [29] | 30,000 | Twitter | Sports, Entertainment, Religion, Crime, Politics, Celebrity, Others | API |
| BHSSP [10] | 20,000 | Bengali Websites | – | Beautiful-Soup Python Library [b] |
| BengFast-Text [9] | 250 Million | Bengali Articles | – | Word Embedding |
| HS-BAN [8] | 5,739 | Facebook | HS, Offensive, Neutral | Manual |
| BD-SHS [5] | 45,000 | Facebook, YouTube | Religious, Political, Gender, Communal, Personal, Others | Semi-Automated Collection and Manual |
| SentNoB [8] | 14,000+ | Blogs, Reviews, Social Media | *Sentiment:* Positive, Negative, Neutral | Manual Annotation |
| DeepHateExplainer [26] | 8,087 | Facebook, YouTube, Newspapers | Political, Personal, Geopolitical, Religious | Manual Annotation |

[a] Facepager: A GUI-based crawler used for fetching data from social media APIs.

[b] Beautiful-Soup: A Python library used for web scraping content from HTML pages of Bengali websites.

[c] SentNoB is a sentiment analysis dataset, included here for its relevance to Bangla text classification tasks.

**Table 3** Dataset 1: NNTI – HS Labeling by Topic [a]

| Bangla Texts (Romanized) | Translation | HS Label | Context |
|---|---|---|---|
| যত্তসব পাপন শালার ফাজলামী!!!!! (Jattosob Papon shalar fazlami!!!!!) | All this nonsense from that Papon bastard!!!!! | HS | Sports |
| পাপন শালা রে রিমান্ডে নেওয়া দরকার (Papon shala re remande neowa dorkar) | That Papon bastard needs to be taken into remand | HS | Sports |
| জারজ হবে এটা কখনও ভাবিনি (Jaroj hobe eta kokhono bhabini) | I never thought he would turn out to be a bastard | HS | Personal |
| শালা লুচ্চা দেখতে পাঠার মত(Shala luchcha dekhte pathar moto) | That scoundrel looks like a goat | HS | Personal |
| তুর মার হেডায় খেলবে সাকিব(Tur mar heday khelbe Shakib) | Shakib will be fucking your mother | HS | Sports |

[a] *Social media comments with category-wise hate annota**tion***

**Table 4** Dataset 2: DeepHateExplainer – Multilabel Dataset Classified into 4 Categories [a]

| Bangla Texts (Romanized) | Translation | HS Label | Context |
|---|---|---|---|
| ৭৫ হামলার আসামী কে এই গডফাদার কালা মন্দির? (75 hamlar asami ke ei godfather kala mondir?) | Who is this godfather Kala Mondir, accused in the '75 attack? | Not HS | Personal |
| রাজাকারদের মন্ত্রী বানানোর দায়ে তার আরও কঠিন শাস্তি দরকার (Rajakarder montri bananor daye tar aro kothin shasti dorkar) | He deserves even harsher punishment for making the Razakars ministers | HS | Political |
| বৃটিশ নাগরিকদের বাংলাদেশ ভ্রমনে সতর্কতা জারী (British nagorikder Bangladesh bhromone sotorkota jari) | Travel advisory issued for British citizens visiting Bangladesh | HS | Geopolitical |
| নয়াদিগন্ত রিপোর্ট করলো হিলারী শপথ নিলেন (Noyadigonto report korlo Hillary shopoth nilen) | Naya Diganta reported that Hillary took the oath | Not HS | Political |
| অসভ্য বর্বর বেহায়া ভোট ডাকাত (Oshobhyo borbor behaya vote dakat) | Uncivilized, barbaric, shameless vote thief | HS | Geopolitical |

[a] *Bengali Dataset for HS detection in socio-political context.*

**Table** 5 Dataset 3: BD-SHS – Annotated for HS and Context [a]

| Bangla Texts (Romanized) | Translation | HS Label | Context |
|---|---|---|---|
| শয়তানের সাথে মিত্রতা করেছে ওরা (Shoytanner sathe mitrota koreche ora) | They have allied with the devil | HS | Religious |
| ওই রাজাকারদের বিচার চাই (Oi rajakarder bichar chai) | We demand justice for those Razakars | HS | Political |
| সব রাজনীতিবিদ চোর (Shob rajnaitibid chor) | All politicians are thieves | HS | Political |
| তুমি খুব ভালো বলছো ভাই (Tumi khub bhalo bolcho bhai) | You are speaking very well, brother | Not HS | General |
| ভালো কাজ করেছো, ধন্যবাদ (Bhalo kaj korecho, dhonnobad) | You did a good job, thank you | Not HS | General |

[a] BD-SHS- Social media comments

### 4.1.2 Merged Datasets

A large, merged dataset was created by combining several benchmark corpora and various subsets. This corpus is the union of the three primary benchmark datasets described above (NNTI, DeepHateExplainer, BD-SHS and additional subsets of SentNoB and BengFastText. Unlike each of the individual datasets used for baseline analysis, this merged dataset was created in the hope that it would more closely simulate more heterogeneous, real-world distributions by pooling content from multiple platforms, annotation schemes, and topics. It encompasses a broad range of HS contexts—political, religious, individual, and geopolitical and is a challenging testbed for evaluating model generalization and robustness.

### 4.1.3 Real World SMP Dataset

To evaluate the generalization limitations of existing Bangla hate speech (HS) detection models, we curated a real-world dataset that captures implicit, culturally dependent, and emoji-laden online discourse. The dataset serves as a diagnostic corpus designed to test whether benchmark-trained models often developed in highly structured conditions can effectively interpret the dynamic, informal, and context-rich environment of social media platforms (SMPs). Unlike large-scale benchmark corpora such as BD-SHS or HS-BAN, which primarily contain explicit and well-structured hate content, this dataset reflects the nuanced and often indirect forms of hostility found in real conversations. It was manually collected from Facebook, Twitter (X), and YouTube, covering a variety of domains such as politics, entertainment, religion, sports, and gender discourse. → Fig. 2 illustrates the distribution of posts across these platforms, with Facebook contributing 45%, Twitter 32%, YouTube 23%, and an additional 8% synthetic samples generated for controlled testing.

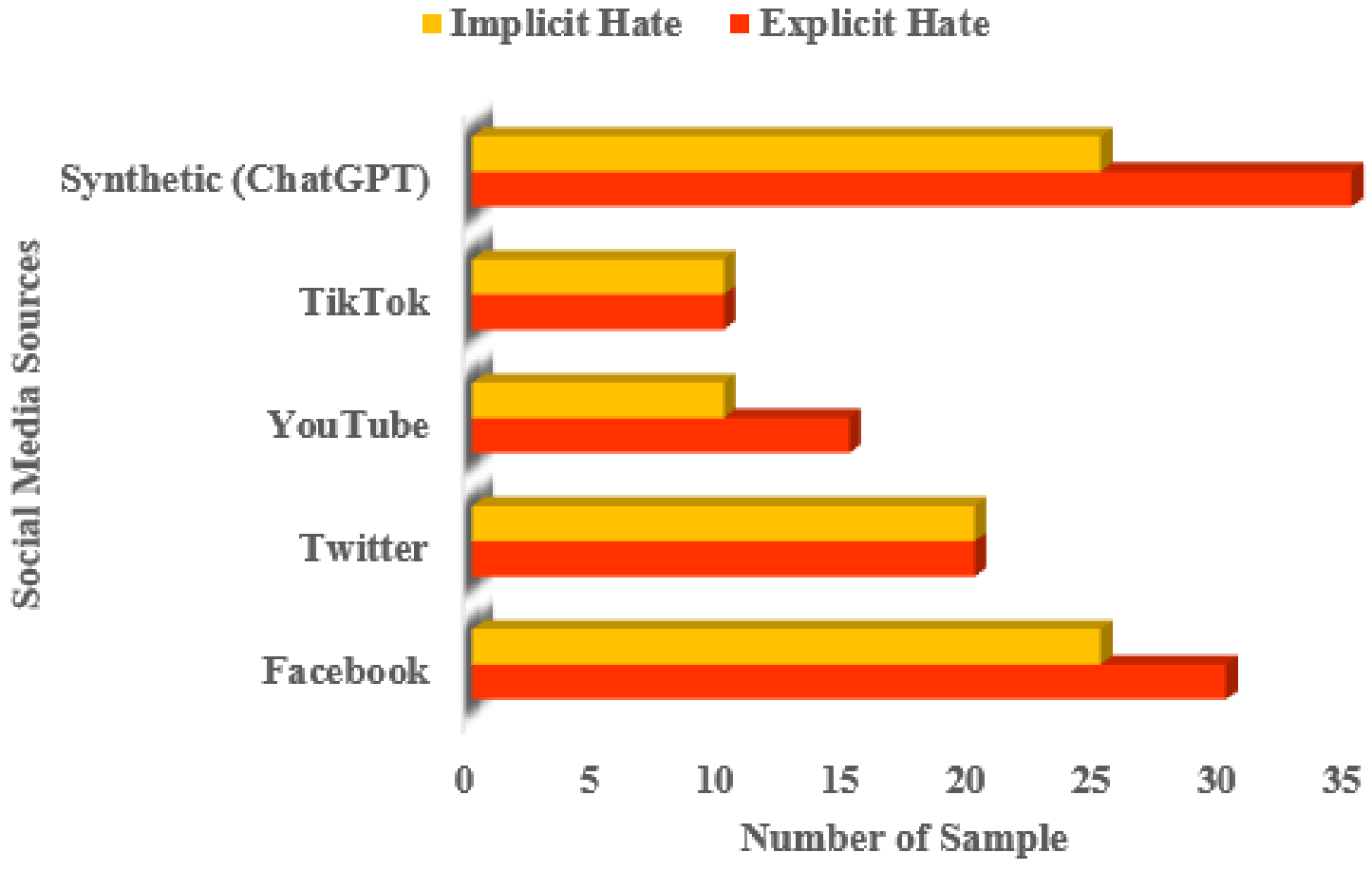

**Fig. 2** Distribution of Manually Collected across different SMPs and Synthetic HSs.

The dataset was annotated following established Bangla HS labeling guidelines (Karim et al., 2021; Romim et al., 2022), with three primary classes: Explicit HS, Implicit HS, and Not HS. To enable cross-comparison with benchmark models, a binary representation was also created, where Explicit and Implicit HS were merged into a single HS category.

However, the fine-grained features were retained for deeper qualitative and error analyses detail in → Sec. 4.3. Each instance includes the following annotation features:

- **Text:** The original Bangla comment/post, possibly code-mixed or emoji-rich.
- **Label:** One of {Explicit HS, Implicit HS, Not HS}.
- **Actual Meaning:** The literal and contextual interpretation of the message.
- **Intended Tone/Category**: Aggressive insult, teasing, political hate, sarcastic trolling, threat, humor, etc.
- **Possible Interpretation:** Cultural or pragmatic explanation of how the text could be understood differently depending on context.

For example,

**Bangla (original):** "তুই অনেক বড় দেশপ্রেমিক! 😂" (*Tui onek boro deshpremik! 😂*)
**Translation:** "You're such a great patriot! 😂"
**Category:** Implicit HS (political sarcasm)
**Interpretation:** The laughing emoji inverts the literal compliment into ridicule, reflecting sarcasm.

Another example illustrates mixed emotion and humor:

**Bangla:** "এই মেয়েটার চেহারা দেখলেই মনে হয় নাটকের ভিলেন 😂🙃" (*Ei meyetar chehara dekhlei mone hoy natoker villain 😂🙃*)
**Translation:** "This girl looks like a TV villain 😂🙃"
**Category:** Implicit HS (teasing, body-shaming)
**Interpretation:** The combination of mocking and upside-down emojis conveys ridicule masked as humor.

These examples demonstrate how implicit cues, emojis, and tone modulations complicate machine interpretation and justify the inclusion of auxiliary semantic and pragmatic features. (more examples are illustrated inside → Table 12, 13)

#### *4.1.3.1* Statistical Insights

The corpus contains approximately 200 Bangla SMP posts or comments, totaling 12,480 tokens (mean = 62.4 tokens per instance, SD = 19.8). Nearly 46% of texts are code-mixed with English, and 48% contain emojis (average = 2.1 per comment). The most frequent emojis are 😂, 🤔, 🙃, and 😡, reflecting humor, sarcasm, irony, and anger respectively affective signals critical for understanding implicit HS.

**Table** 6 Lebel Distribution & Intended tone categories

| Label Type | Count | Percentage | Dominant Tone Categories |
|---|---|---|---|
| Explicit HS | 70 | 35% | Aggressive insult, threat, religious slur |
| Implicit HS | 60 | 30% | Political sarcasm, teasing, trolling |
| Not HS | 70 | 35% | Neutral, supportive humor, informational comment |
| Total | 200 | 100% | — |

**Table 7** Platform & Domain Composition

| Platform | Share (%) | Topical Domain | Avg. Comment Length |
|---|---|---|---|
| Facebook | 45 | Political debates, celebrity criticism | 68 |
| Twitter (X) | 32 | News reaction, political satire | 59 |
| YouTube | 23 | Entertainment, sports commentary | 52 |
| Synthetic (Augmented) | 8 | Sarcastic paraphrases, coded hate | 61 |

Facebook samples contained the highest density of explicit hate, while Twitter contributed most implicit and sarcastic content. YouTube comments were typically shorter and humor-based. Synthetic samples (≈ 8%) were generated through controlled paraphrasing and context-preserving rewording to balance underrepresented implicit categories. All synthetic entries were manually reviewed by annotators before inclusion to ensure semantic validity. Emoji analysis revealed that translating emojis into textual sentiment improved BanglaBERT's F1-score from 0.63 to 0.75, highlighting their affective significance in computational detection. The dataset's token irregularities, phonetic spellings, and humor markers accurately represent informal Bangla SMP language, crucial for evaluating model robustness beyond benchmark constraints.

**Table** 8 Summary Statistics

| ***Metric*** | ***Value*** |
|---|---|
| Total Tokens | 12,480 |
| Mean Tokens per Comment | 62.4 |
| Code-Mixed Comments | 46% |
| Emoji-Containing Comments | 48% |
| Avg. Emojis per Comment | 2.1 |
| Distinct Emojis Used | 137 |
| Avg. Words per Explicit HS | 8.7 |
| Avg. Words per Implicit HS | 11.2 |
| Inter-Annotator Agreement (κ) | 0.81 |
| Synthetic Samples | 15 (7.5%) |
| Collection Period | March–August 2024 |

The dataset thus functions as a high-context diagnostic corpus, not a benchmark, and enables systematic exploration of model misclassification causes under diverse linguistic and emotional conditions.

#### *4.1.3.2* Ethical and Data Governance Considerations

All data were collected from publicly accessible social media content while adhering to rigorous ethical and privacy standards. Usernames, links, and any identifiable metadata were removed or replaced with anonymized tokens (*[user]*, *[page]*). Posts containing personal identifiers or sensitive private details were excluded during preprocessing. The dataset was developed in compliance with GDPR principles and the Digital Security Act (DSA) 2018 for digital content handling. Annotators received ethics briefings and provided informed consent prior to participation. They were trained to handle offensive material responsibly, and psychological support procedures were made available in case of distress during annotation. To preserve free speech context while avoiding harm, borderline or ambiguous samples particularly political satire or teasing were intentionally retained. This helps analyze over-policing tendencies in HS detection systems and supports balanced moderation strategies. Due to the sensitive nature of the data, full release is restricted. However, anonymized samples, annotation schema, and metadata summaries will be available upon reasonable academic request. This approach aligns with *Information Processing & Management's* guidelines for ethical data governance, transparency, and reproducibility.

## *4.2 Preprocessing Steps*

As Bangla social media text is noisy and has an emotional, informal taste, a sequence of preprocessing methods were applied to sanitize it: stop-word removal with a more comprehensive Bangla stop-word list, emoji translation via a custom dictionary and use of the bnemo library[1] for preservation of emotional semantics, hashtag segmentation and removal of

[1] bnemo — bnemo latest documentation. https://bnemo.readthedocs.io/.

uninformative tokens, Sentence tokenization to be compatible with transformer-based models, and general normalization that involves punctuation removal, stemming, and script normalization. However, the specific preprocessing operations varied depending on the type of embedding used FastText or BanglaBERT to ensure that each model received optimally prepared input without disrupting its native tokenization mechanisms.

1. **For FastText-based models (FastText + CNN, FastText + LSTM, FastText + BiLSTM):** Extensive preprocessing was applied to enhance word-level embedding quality. These operations included stop-word removal using an enriched Bangla stop-word list (≈440 entries), stemming, punctuation and number removal, normalization of orthographic variants, and removal of infrequent or non-Bangla tokens. Such preprocessing is crucial for static embeddings like FastText, which lack contextual subword tokenization and rely heavily on clean lexical units for representation.
2. **For Transformer-based models (BanglaBERT, BanglaBERT + CNN, BanglaBERT + BiLSTM):** Minimal preprocessing was used to preserve contextual integrity, since BanglaBERT employs its own SentencePiece tokenizer trained on large-scale Bangla text. We did not perform stop-word removal, stemming, or lemmatization for these models. Only lightweight cleaning steps—such as removing URLs, hashtags, HTML entities, user mentions, and redundant whitespace—were applied to prevent tokenization conflicts while maintaining the linguistic and contextual richness of the text.

The overall text preprocessing steps are described in details below:

#### 4.2.1 Text Cleaning

This is the initial step, which is the foundation of data refinement. In this process, unwanted things may have been attached while they were being obtained. Both the raw data and existing datasets were pre-processed. Except for some existing datasets, this step was used on other datasets, especially on the raw data that were obtained by scraping comments from different SMPs.

#### 4.2.2 Tokenization

After text cleaning, sentence piece and word piece tokenization were applied to the Bangla text-based datasets. Tokenization is the method of splitting the text into its basic constituents, i.e., words or tokens. There is a library support for tokenizing Bangla text. For example, in word piece tokenization, the sentence “এদেশে কোন রাজাকারের ঠাই নাই ।' after tokenization is split as ['এদেশে', 'কোন', 'রাজাকারের', 'ঠাই', 'নাই', '।']

#### 4.2.3 Hashtag Segmentation

Hashtags are usually employed by social media platform (SMP) users to create keywords or stress the significance of a topic. However, they do not inherit any meaningful importance. Some researchers have removed hashtags and used spaces instead, maintaining the keywords they hold for future use. This paper (Jahan et al., 2022) illustrated that the word 'হতাশ' has been translated into a depression by replacing the hashtag with a space, i.e., 'হতাশ' (depression).

#### 4.2.4 Stop word Removal

Stop word removal in Bengali occurs through the process of eliminating stop words that do not carry significant meaning or aid in the understanding of the text. Stop word elimination from Bangla text is a preprocessing method utilized in text classification and text analysis. The stop word list of the Bangla language available is further enriched by existing research (Mridha et al., 2021). The list consists of 440 words with several prepositions, quantifiers, conjunctions, and linkers. By eliminating stopwords from the text because they generally do not have any meaningful content, it enhances the dataset's readability and focuses on semantically meaningful content.

### 4.2.5 Stemming

Stemming in Bengali speech processing is used to transform inflected words into their basic stems, bases, or root forms. This process helps reduce dimensionality and provides more consistent Bengali tokens. For example, the word "বাংলাদেশী" (Bangladeshi) is transformed into "বাংলাদেশ" (Bangladesh).

### 4.2.6 Emoji And Emoticon Conversion

Emojis and emoticons play crucial roles in detecting offensive content. The heterogeneity of data types, such as emojis, emoticons, and words, allows text processing to advance. For emoji translation, comments containing emojis are translated by replacing them with similar Bengali sentences. To understand the effect of emojis in HS detection, two approaches were compared: (1) erasing all the emojis, and (2) translating the emojis into Bangla forms of text. This allowed us to study whether preserving the contextual and emotional content conveyed by emojis enhances model performance. Our findings (→ Sec. 5.3.3) are that emoji translation enhances classification accuracy, especially in detecting implicit or sentiment-hate, and vindicate the significance of culture-aware preprocessing in Bangla NLP. The emojis were replaced with similar Bangla text conveying the same emotion with the Python bnemo library accessible for public use.

### 4.2.7 Miscellaneous

PoS tagging, removal of infrequent words, removal of punctuation, removal of non-Bengali words and symbols, numbers of words to words, elimination of accent marks, and whitespace are some of the other techniques also used for preprocessing Bangla text. All these preprocessing stages combine to assist in the generation of high-quality and standard datasets, which will further benefit the use of NLP algorithms or models for downstream tasks such as text classification for HS.

## *4.3 Models*

To evaluate the effectiveness of various sequence modeling and text representation methods in Bangla HS detection, six diverse architectures were compared. The models are categorized into two main categories based on their embedding backbone: transformer-based (BanglaBERT variants) and traditional word embedding-based (FastText variants). Each model was trained on the joint set and then evaluated on three alternative test sets: benchmark corpora, the joint set, and an external sample of real-world data containing both implicit and explicit HSs.

### 4.3.1 Transformer-Based Models

#### *4.3.1.1* BanglaBERT (Base)

BanglaBERT is a transformer-based model built using the architecture of BERT, except that it has been pretrained on 27.5 GB of Bangla data (dubbed 'Bangla2B+') by crawling 110 popular Bangla sites. Its performance on Natural Language Understanding (NLU) tasks was evaluated using datasets on natural language inference and question answering, and benchmarked across four diverse NLU tasks (text classification, sequence labelling, span prediction). BanglaBERT achieved state-of-the-art results, outperforming multilingual and monolingual models [25]. The model comprises 12 encoder layers, 12 attention heads, a hidden size of 768, and approximately 110 million parameters.

#### *4.3.1.1.1 Embedding Layer*

Given a tokenized input sequence $X = \{x_1, x_2, \dots, x_T\}$ of length $T$, BanglaBERT computes an embedding $e_i \in R^d$ for each token $x_i$ as:

$$\mathbf{e_i} = E_t(x_i) + E_s(x_i) + E_p(i) \quad (1)$$

Where $E_t$ is the token embedding lookup matrix, $E_s$ is the segment embedding (used for sentence-pair tasks), and $E_p$ is the positional embedding. The final input embedding matrix for the sequence is as follows.

$$\boldsymbol{E} = [\boldsymbol{e_1}, \boldsymbol{e_2}, \dots, \boldsymbol{e_T}] \in R^{T\times d} \quad (2)$$

#### *4.3.1.1.2 Positional Encoding*

To retain order information, BanglaBERT applies sinusoidal positional encodings as proposed in previous research (Vaswani et al., 2023). The positional encoding vectors are defined as:

$$\mathrm{PE}_{(pos,2i)} = \sin\left(\frac{pos}{10000^{2i/d}}\right) \quad (3)$$

$$\mathrm{PE}_{(pos,2i+1)} = \cos\left(\frac{pos}{10000^{2i/d}}\right) \quad (4)$$

These are added to the token embeddings to form the final model input:

$$X = \boldsymbol{E} + \mathrm{PE} \quad (5)$$

#### *4.3.1.1.3 Transformer Encoder and Attention*

BanglaBERT uses 12 stacked Transformer encoder layers. Each includes a multi-head self-attention mechanism followed by a feed-forward network. The scaled dot-product attention is defined as:

$$\mathrm{Attention}(Q, K, V) = \mathrm{softmax}\left(\frac{QK^{\top}}{\sqrt{d_k}}\right) V \quad (6)$$

where,

$$Q = XW^Q, \quad K = XW^K, \quad V = XW^V \quad (7)$$

and $W^Q, W^K, W^V \in R^{d\times d_k}$ are learned projection matrices. Multi-head attention is computed by projecting $Q, K, V,$ and into $h$ heads, computing scaled dot-product attention for each, and concatenating the outputs:

$$\mathrm{MultiHead}(Q, K, V) = \mathrm{Concat}(\mathrm{head}_1, \dots, \mathrm{head}_h)W^O \quad (8)$$

$$\mathrm{head}_i = \mathrm{Attention}\left(QW_i^Q, KW_i^K, VW_i^V\right) \quad (9)$$

#### *4.3.1.1.4 Feed-Forward and Layer Normalization*

Each encoder layer contains a position-wise feed-forward network:

$$\mathrm{FFN}(x) = \max(0, xW_1 + b_1)\, W_2 + b_2 \quad (10)$$

This is followed by a residual connection and layer normalization:

$$\mathrm{LayerNorm}\left(x + \mathrm{Dropout}\big(\mathrm{Sublayer}(x)\big)\right) \quad (11)$$

#### *4.3.1.1.5 Pretraining with ELECTRA*

BanglaBERT was pretrained using the ELECTRA framework (Clark et al., 2020), in which a discriminator is trained to predict whether a generator has replaced each token in an input sequence. Unlike the traditional Masked Language Modeling (MLM) approach, ELECTRA trains on all tokens, thereby improving sample efficiency. During pretraining, a generator first predicts masked tokens using standard MLM; the masked tokens are then replaced with samples from the generator's output, and finally, the discriminator is trained to distinguish which tokens were replaced. The pretraining was conducted with a batch size of 256 for 2.5 million steps on TPUs using the Adam optimizer with a learning rate of $2 \times 10^{-4}$ and a warmup phase of 10,000 steps. The vocabulary was generated using Byte-Pair Encoding (BPE), and no lemmatization or stemming was applied.

#### *4.3.1.2* BanglaBERT + CNN

To explore the potential benefits of combining contextual embeddings with local pattern extraction, a hybrid architecture was implemented by stacking CNN layers on top of BanglaBERT embeddings. This model aims to capture local phrase-level features—such as repeated abusive expressions or short syntactic patterns—that may be overlooked by the global attention mechanism in standard transformers (Belal & others, 2023; Mridha et al., 2021).

Given an input sequence $\mathrm{X} = \{\mathrm{x}_1, \mathrm{x}_2, \ldots, \mathrm{x}_\mathrm{T}\}$, the contextualized token representations generated by the final transformer layer of BanglaBERT are denoted as:

$$H = [h_1, h_2, \ldots, h_T] \in R^{T\times 768} \quad (12)$$

To extract discriminative local features, multiple 1D convolutional filters were applied with kernel sizes $\mathrm{k} \in \{2, 3, 4\}$ and 100 filters per size were applied. Each convolutional operation is followed by a ReLU activation function:

$$f_i = \mathrm{ReLU}(W_i * H + b_i), \quad i = 1,2,3 \quad (13)$$

where $W_i \in R^{k\times 768}$ denotes the convolution kernel and where $*$ represents a 1D convolution. Each feature map $\mathrm{f_i}$ is then downsampled using global max pooling to retain the most salient information. The pooled outputs are concatenated into a single fixed-length vector:

$$c = [\mathrm{maxpool}(f_2),\ \mathrm{maxpool}(f_3),\ \mathrm{maxpool}(f_4)] \in R^{300} \quad (14)$$

A dropout layer with a rate of 0.5 is applied to $c$ to mitigate overfitting. Finally, the regularized representation is passed through a fully connected layer followed by softmax for binary classification:

$$\hat{y} = \mathrm{softmax}(W_c \cdot c + b_c) \quad (15)$$

#### *4.3.1.3* BanglaBERT + Bi-LSTM

Another hybrid architecture also was constructed by feeding BanglaBERT embeddings into a Bidirectional Long Short-Term Memory (Bi-LSTM) network. This design aims to capture long-range dependencies and sequential patterns in the text, which are crucial for understanding nuanced expressions and contextual cues in Bangla HS classification. The BanglaBERT embeddings are passed to a Bi-LSTM layer with a hidden size of 128, which processes the sequence in both forward and backward directions to generate rich contextualized representations:

$$\overrightarrow{h_t} = \mathrm{LSTM}_{\mathrm{forward}}\left(h_t, \overrightarrow{h_{t-1}}\right), \quad \overleftarrow{h_t} = \mathrm{LSTM}_{\mathrm{backward}}\left(h_t, \overleftarrow{h_{t+1}}\right) \quad (16)$$

The forward and backward hidden states are concatenated at each time step:

$$\mathrm{h}'_\mathrm{t} = \left[\overrightarrow{\mathrm{h}_\mathrm{t}}; \overleftarrow{\mathrm{h}_\mathrm{t}}\right] \in \mathrm{R}^{256} \quad (17)$$

To obtain a fixed-length sentence representation, a max-pooling operation is applied over all time steps:

$$c = \max_{t=1}^{T} h'_t \in R^{256} \quad (18)$$

A dropout layer with a rate of 0.3 is applied to **c** to reduce overfitting. The BiLSTM is trained using the Adam optimizer with a learning rate of 0.001. The model is trained for 10 epochs with a batch size of 32, minimizing the binary cross-entropy loss. Finally, the regularized vector is passed through a fully connected layer followed by a softmax activation for binary classification (Equation 15).

Both hybrid architectures effectively combine BanglaBERT's rich contextual embeddings with the sequential learning capabilities of BiLSTM and the CNN's strength in capturing position-invariant local features, enabling the model to capture both semantic and syntactic nuances present in Bangla HS.

## 4.3.2 FastText-Based Models

### *4.3.2.1* FastText + CNN

To exploit both subword-level semantic richness and local n-gram-based feature extraction, a hybrid architecture that combines FastText embeddings with a Convolutional Neural Network (CNN) classifier was applied. FastText represents each word as a composition of character $n$-grams, enabling the model to effectively handle rare, misspelled, or morphologically rich words—common in noisy Bangla social media text.

#### *4.3.2.1.1FastText Embedding Process*

Given a tokenized sentence $X = \{x_1, x_2, \ldots, x_T\}$ of length T, each word $x_t$ is decomposed into a set of character-levels $\mathrm{n-grams}$:

$$G(x_t) = \{g_1, g_2, \ldots, g_m\} \tag{19}$$

Each word's embedding $\mathrm{e_t} \in \mathrm{R^d}$ is computed as the average of its constituent n-gram vectors:

$$\mathrm{E_t} = \frac{1}{|\mathrm{G(x_t)}|} \sum_{\mathrm{g \in G(x_t)}} \mathrm{v_g} \tag{20}$$

This results in an embedding matrix for the sentence:

$$E = [e_1, e_2, \ldots, e_T] \in R^{T \times d} \tag{21}$$

where $d = 300$ is the FastText embedding dimension.

#### *4.3.2.1.2 CNN-Based Feature Extraction*

To extract local patterns such as offensive bigrams and trigrams, 1D convolutional filters of widths $k \in \{2, 3, 4\}$ are applied over the embedding matrix. Each filter $W_i \in R^{k \times d}$ produces a feature map:

$$f_i = \mathrm{ReLU}(W_i * E + b_i) \tag{22}$$

where $*$ denotes the convolution operation and ReLU is the activation function. 100 filters for each kernel size were used. Each feature map $f_i$ is reduced via max-pooling over time:

$$\mathrm{C_i = max(f_I)}, \quad \mathrm{c_i \in R^{100}} \tag{23}$$

The pooled vectors are concatenated to form the final representation:

$$\mathrm{C = [c_2;\ c_3;\ c_4] \in R^{300}} \tag{24}$$

After applying a dropout rate of 0.5, the final classification is performed using a softmax layer:

$$\hat{y} = \mathrm{softmax}(W_c \cdot \mathrm{Dropout}(c) + b_c) \tag{25}$$

#### *4.3.2.1.3 Training Configuration*

Standard hyperparameter settings commonly used in HS classification for Bangla (Romim et al., 2022) were adopted, including an Adam optimizer with a learning rate of $1 \times 10^{-3}$, batch size of 32, and early stopping based on validation loss.

#### *4.3.2.2* FastText +LSTM

This model integrates FastText embeddings with a unidirectional Long Short-Term Memory (LSTM) network to capture sequential dependencies in Bangla HS. The embedded sequence is passed through an LSTM layer, where the hidden state at each time step t is computed as

$$h_t = \text{LSTM}(e_t, h_{t-1}) \tag{26}$$

with $e_t \in R^d$ is the input word vector at time t, and where $h_t \in R^h$ the hidden state. The final hidden state $h_T$ summarizes the sequence and is passed through a dropout layer (rate = 0.3), followed by a softmax classifier:

$$\hat{y} = \text{softmax}(W_o \cdot \text{Dropout}(h_T) + b_o) \tag{27}$$

where $W_o \in R^{2\times h}$ and $b_o \in R^2$ are the output layer weights and bias respectively. This architecture is effective in modeling the long-range context and narrative-style abuse patterns that are common in implicit Bangla HS.

#### *4.3.2.3* FastText + Bi-LSTM

This model combines FastText embeddings with a bidirectional Long Short-Term Memory (BiLSTM) network to capture both past and future contexts in Bangla HS sequences. BiLSTMs have shown robust performance in Bangla NLP tasks involving social media text.

Given a sequence of static embeddings $E = [e_1, \dots, e_T]$, the BiLSTM computes forward and backward hidden states:

$$\overrightarrow{h_t} = \text{LSTM}_{\text{fwd}}\left(e_t, \overrightarrow{h_{t-1}}\right), \quad \overleftarrow{h_t} = \text{LSTM}_{\text{bwd}}\left(e_t, \overleftarrow{h_{t+1}}\right) \tag{28}$$

$$h_t = \left[\overrightarrow{h_t}; \overleftarrow{h_t}\right] \in R^{2h} \tag{29}$$

The concatenated final hidden state $h_T = \left[\overrightarrow{h_T}; \overleftarrow{h_1}\right]$ is passed through a dropout (rate = 0.3) and a SoftMax classifier:

$$\hat{y} = \text{softmax}(W_o \cdot \text{Dropout}(h_T) + b_o) \tag{30}$$

where $W_o \in R^{2\times 2h}$ and $b_o \in R^2$. The number of hidden units used, $h = 128$, the Adam optimizer with a learning rate of $1 \times 10^{-3}$, batch size 32, and early stopping—hyperparameters aligned with prior works on Bangla HS detection.

## *4.4 Evaluation Strategy*

To ensure a thorough and consistent assessment of model performance, a unified evaluation strategy was employed across all the models. Each model was trained using the merged dataset, which combines multiple publicly available HS corpora to ensure linguistic and contextual diversity. The evaluation was then conducted separately on three distinct test sets: (i) benchmark datasets commonly used in the existing literature, (ii) a subset of the merged dataset held out for testing, and (iii) a manually annotated real-world dataset consisting of informal, emoji-rich, and implicit HS expressions collected from Bangla social media platforms. For performance measurement, standard classification metrics: accuracy, precision, recall, and F1-score were used. These metrics were computed both on a per-class basis and as macro-averaged and weighted-average aggregates. Accuracy provides a general measure of correct predictions, whereas precision and recall reflect the model's ability to handle false positives and false negatives, respectively. The F1-score, which is the harmonic mean of precision and recall, was emphasized due to the class imbalance present in the real-world dataset. The formulas for the main evaluation metrics are as follows:

$$\text{Precision} = \frac{\text{TP}}{\text{TP} + \text{FP}} \tag{31}$$

$$\text{Recall} = \frac{\text{TP}}{\text{TP} + \text{FN}} \tag{32}$$

$$F_{1\text{-score}} = 2 \times \frac{\text{Precision} \times \text{Recall}}{\text{Precision} + \text{Recall}} \tag{33}$$

All the experiments were repeated three times using different random seeds, and the final reported metrics represent the average across these runs. The results of these evaluations are presented and discussed in the subsequent sections.

# 5 Result

This section provides quantitative and qualitative findings of model performance on benchmark, merged, and real-world datasets. All the findings collectively provide inferences about the strengths, weaknesses, as well as real-world implications of using HS detection systems in Bangla social media settings.

**Table** 9: Hyperparameters used for Bangla HS Detection across Different Models

| Hyperparameter | Bangla-BERT Base | Bangla-BERT + CNN | Bangla-BERT + Bi-LSTM | FastText + CNN | FastText + LSTM | FastText + Bi-LSTM |
|---|---|---|---|---|---|---|
| Embedding Dimension | 768 | 768 | 768 | 300 | 300 | 300 |
| Number of Layers | 12 | 12 | 12 | N/A | N/A | N/A |
| Attention Heads | 12 | 12 | 12 | N/A | N/A | N/A |
| Hidden Size | 768 | 768 | 768 | 300 | 300 | 300 |
| Parameters | 110M | 110M | 110M | N/A | N/A | N/A |
| Batch Size | 256 | 32 | 32 | 32 | 32 | 32 |
| Epochs | N/A | N/A | 10 | N/A | N/A | 10 |
| Optimizer | Adam | Adam | Adam | Adam | Adam | Adam |
| Learning Rate | $2\times10^{-4}$ | $1\times10^{-3}$ | $1\times10^{-3}$ | $1\times10^{-3}$ | $1\times10^{-3}$ | $1\times10^{-3}$ |
| Dropout Rate | N/A | 0.5 | 0.3 | 0.5 | 0.3 | 0.3 |
| Max Pooling | N/A | Yes | Yes | Yes | Yes | Yes |
| Filter Sizes | N/A | {2,3,4} | {2,3,4} | {2,3,4} | N/A | N/A |
| ReLU Activation | Yes | Yes | Yes | Yes | Yes | Yes |
| LSTM Hidden Size | N/A | N/A | 128 | N/A | 128 | 128 |
| Optimizer Settings | $2\times10^{-4}$ | $1\times10^{-3}$ | $1\times10^{-3}$ | $1\times10^{-3}$ | $1\times10^{-3}$ | $1\times10^{-3}$ |

N/A indicates the hyperparameters do not apply to the corresponding model architecture

## 5.1 Quantitative Results

To evaluate the effectiveness and generalization capability of Bangla HS detection models, six architectures were assessed across three experimental settings: (i) three benchmark datasets—NNTI, DeepHateExplainer, and BD-SHS, (ii) a merged dataset created by combining several existing corpora, and (iii) a manually curated external dataset containing real-world social media content with explicit and implicit HS. All the models were trained on the merged dataset to ensure comparable learning exposure and then tested on each evaluation set individually. The primary metric used for performance comparison was the **macro-averaged F1-score**, which provides a balanced evaluation across imbalanced class distributions. Additional metrics including accuracy, precision, and recall were also considered for a comprehensive understanding of the models' classification performance.

**Table 10**: Model performance on benchmark datasets

| Model | NNTI (F1-score) | DeepHateExplainer (F1-score) | BD-SHS (F1-score) |
|---|---|---|---|
| FastText + CNN | 0.82 | 0.67 | 0.84 |
| FastText + BiLSTM | 0.85 | 0.69 | 0.89 |
| FastText + LSTM | 0.83 | 0.66 | 0.87 |
| BanglaBERT | **0.88** | **0.73** | **0.91** |
| BanglaBERT + CNN | 0.87 | 0.71 | 0.88 |
| BanglaBERT + BiLSTM | 0.86 | 0.70 | 0.86 |

On the benchmark datasets, transformer-based models outperformed their RNN and CNN-based counterparts. BanglaBERT achieved the highest F1-scores across all three benchmarks, highlighting the advantage of deep contextual representation in classifying Bangla HS. Among the lightweight architectures, FastText + BiLSTM yielded the most competitive results, particularly for BD-SHS. The performance on the merged dataset is summarized in → Table 8. The merged dataset introduced greater topical and linguistic diversity, mimicking varied online discourse. BanglaBERT-based models continued to lead, although the performance gap has narrowed slightly due to more complex data patterns. Finally, → Table 9 presents the model performance on the manually annotated external dataset containing both explicit and implicit HS sourced from Bangla social media. All models experienced notable decreases in accuracy and F1-scores, particularly for implicit HS examples characterized by sarcasm, cultural innuendos, and emoji-laden content. BanglaBERT remained the most robust model, although performance degradation was still evident on complex samples. In summary, the quantitative results reinforce the strength of pretrained transformer models, particularly BanglaBERT, in handling structured and benchmark data.

**Table** 11: Model performance on the merged dataset

| Model | Accuracy | F1-score |
|---|---|---|
| BanglaBERT | 91.4% | 0.91 |
| BanglaBERT + CNN | 90.1% | 0.89 |
| BanglaBERT + BiLSTM | 89.6% | 0.88 |
| FastText + BiLSTM | 89.1% | 0.88 |
| FastText + CNN | 85.5% | 0.85 |
| FastText + LSTM | 84.3% | 0.83 |

**Table 12**: Model performance on the external (real-world) SMP dataset

| Model | Explicit HS | Implicit HS | Combined Accuracy |
|---|---|---|---|
| BanglaBERT | 88.2% | 63.4% | 75.3% |
| BanglaBERT + CNN | 85.1% | 59.6% | 72.4% |
| BanglaBERT + BiLSTM | 83.8% | 56.3% | 70.9% |
| FastText + BiLSTM | 82.6% | 58.1% | 70.1% |
| FastText + CNN | 78.0% | 51.2% | 64.6% |
| FastText + LSTM | 75.5% | 49.4% | 62.0% |

However, real-world deployment still presents considerable challenges, especially for implicit HS. These results highlight the necessity for context-aware, culturally informed, and emoji-sensitive Bangla HS detection systems.

### 5.1.1 Confusion Matrix and ROC Curve

Confusion matrices and Receiver Operating Characteristic (ROC) curves were analysed on test sets of benchmark and real-world instances. This was performed to determine the accuracy with which the models can distinguish hateful from non-hateful content, particularly in difficult instances such as implicit HS. → Fig. 3 shows the confusion matrices of three prototype models—BanglaBERT, BanglaBERT + CNN, and BanglaBERT + Bi-LSTM—on the external real-world dataset. All the models produced high true positive rates for overt HS, correctly identifying offensive or overtly aggressive content. However, in the implicit cases, many misclassifications were observed, especially when sarcasm, humor, and emojis were involved. Among them, BanglaBERT achieved the most equitable performance with a higher true negative rate for implicit content and fewer overall false positives. BanglaBERT + CNN and BanglaBERT sometimes confused non-hateful but politically charged or sarcastic content as hate, a source of over-policing issues. ROC curve ( → Fig. 4) analysis also confirms this. BanglaBERT performed the highest AUC (~0.97) on the test samples, but its accuracy dropped to ~0.62 on the actual samples, suggesting difficulty in handling sentiment ambiguity. All these results confirm that the models perform well for overt HS but lose their effectiveness significantly under context-dependent or vague situations.

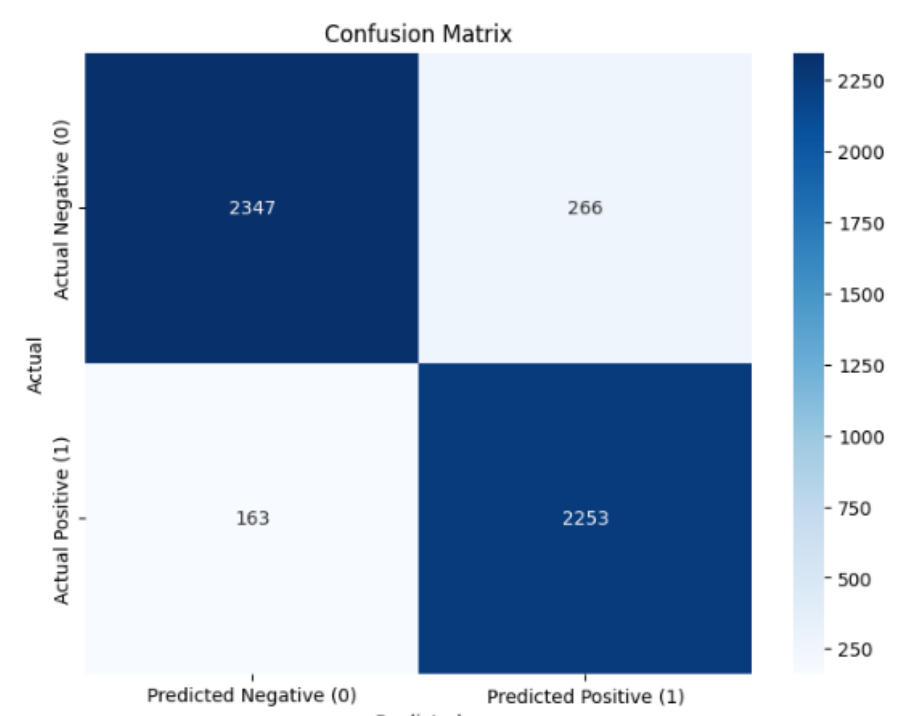

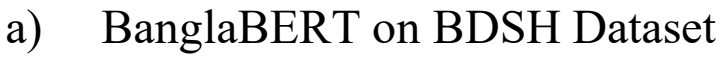
a) BanglaBERT on BDSH Dataset

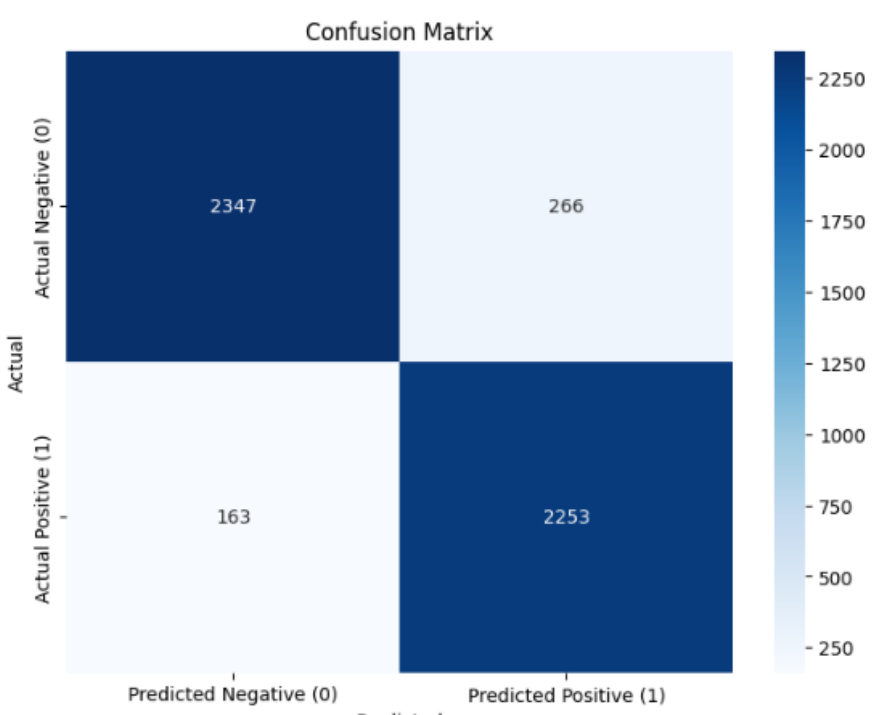

b) BanglaBERT on NNTI Dataset

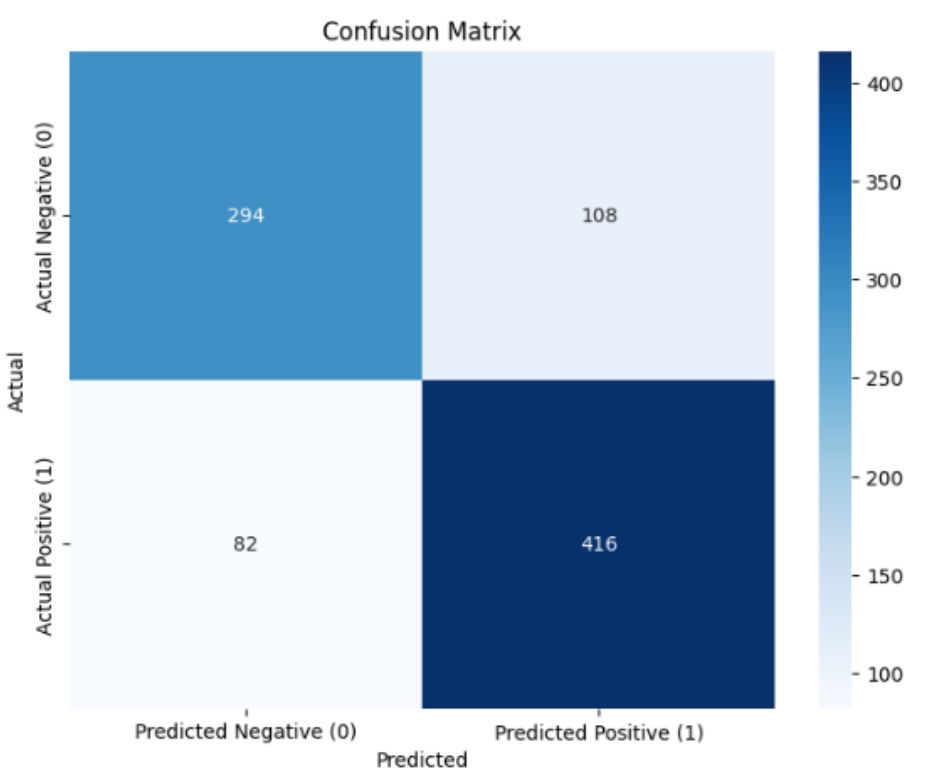

c) BanglaBERT on SentNoBD Dataset

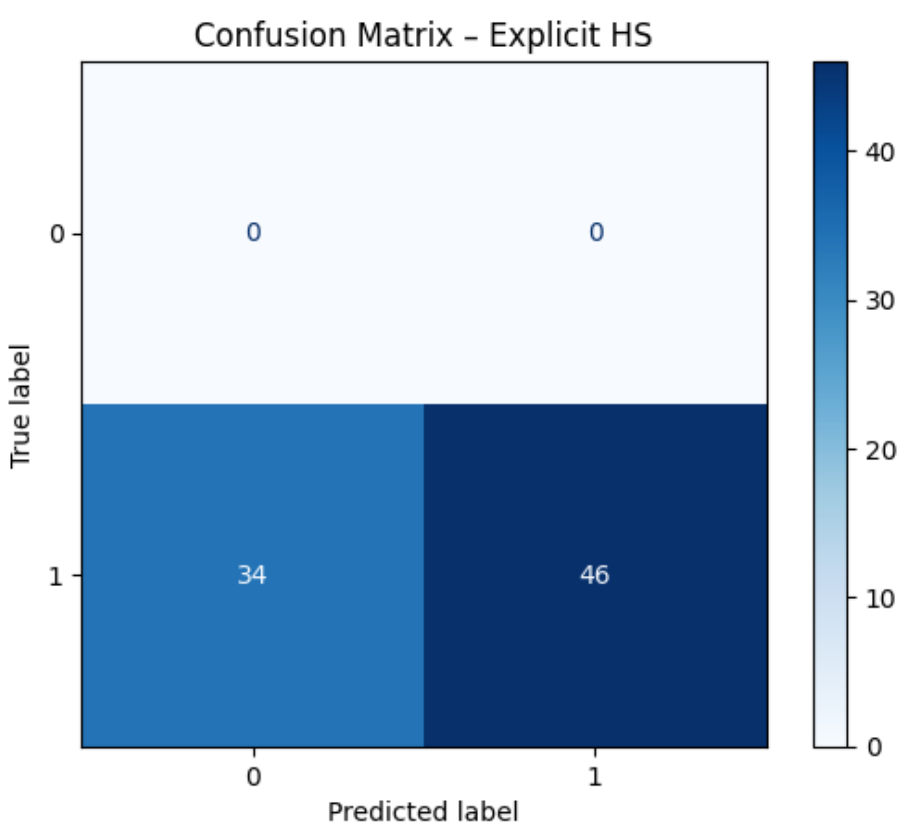

d) BanglaBert (BDSHS) on explicit HS

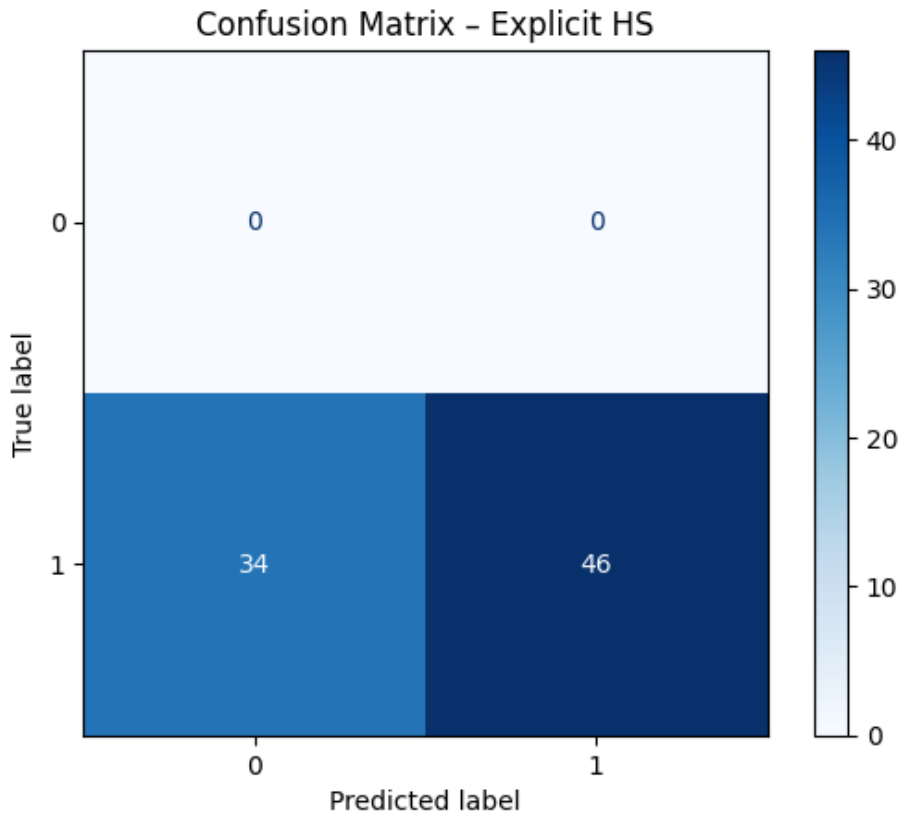

e) BanglaBERT (NNTI) on explicit HS

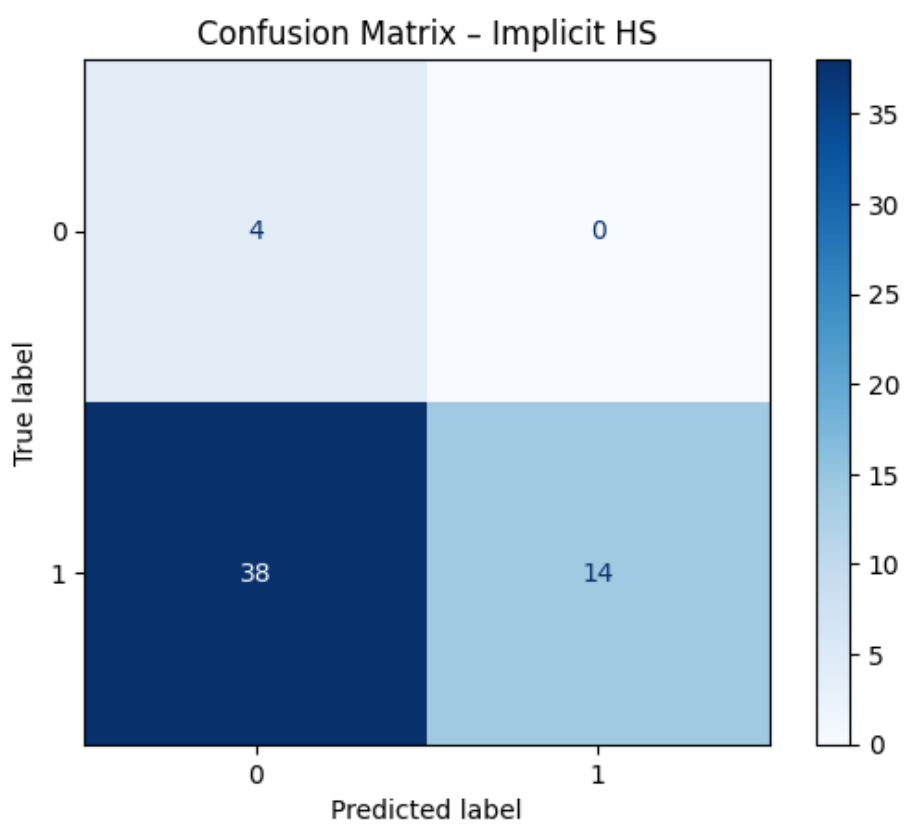

f) BanglaBERT (NNTI) on Implicit HS

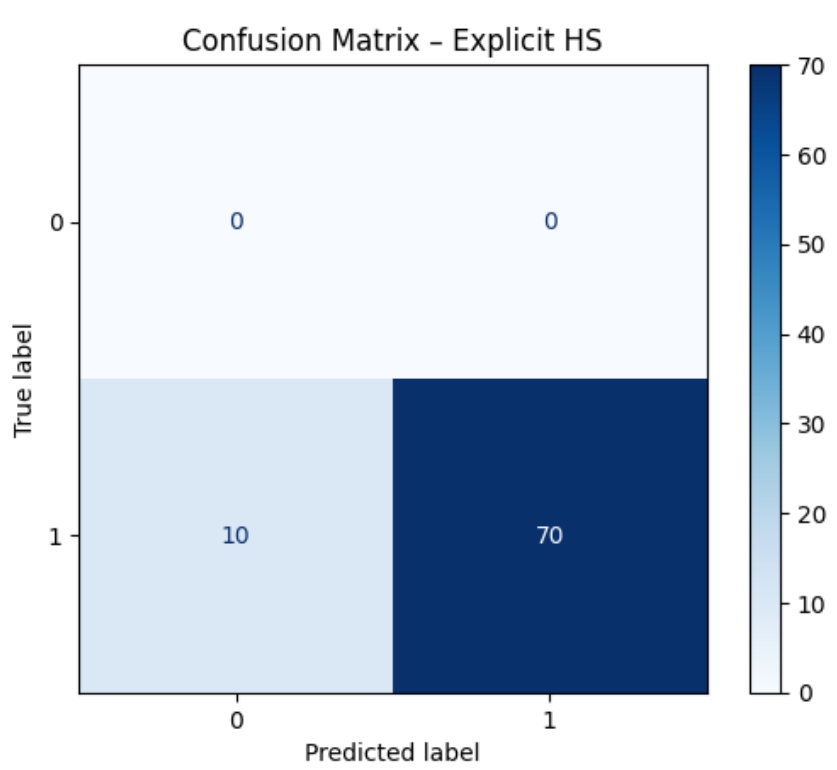

g) BanglaBERT (SentNoBD) on Explicit HS

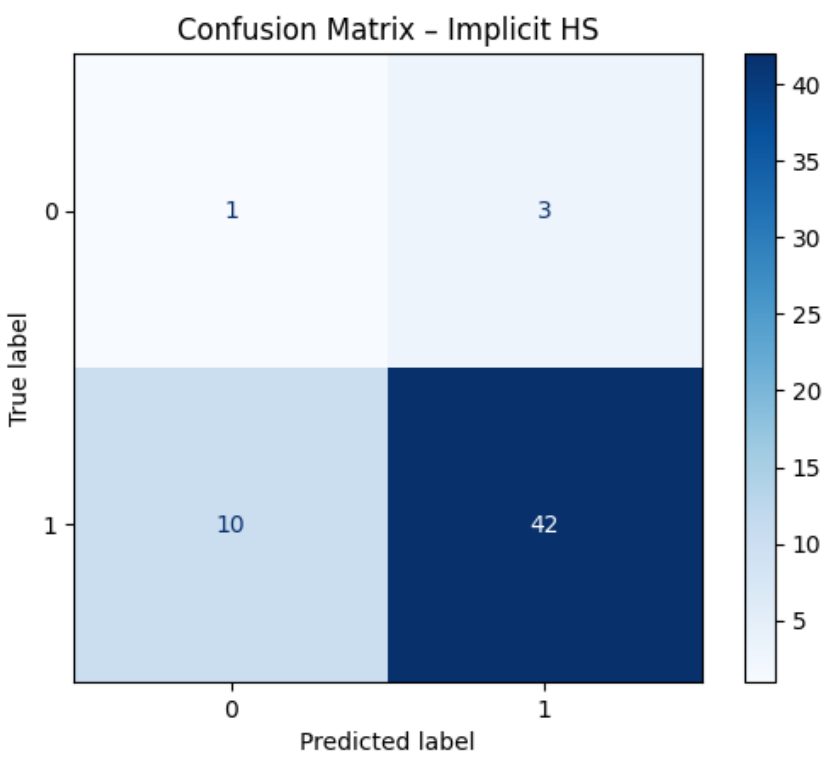

h) BanglaBERT (SentNoBD) on Implicit HS

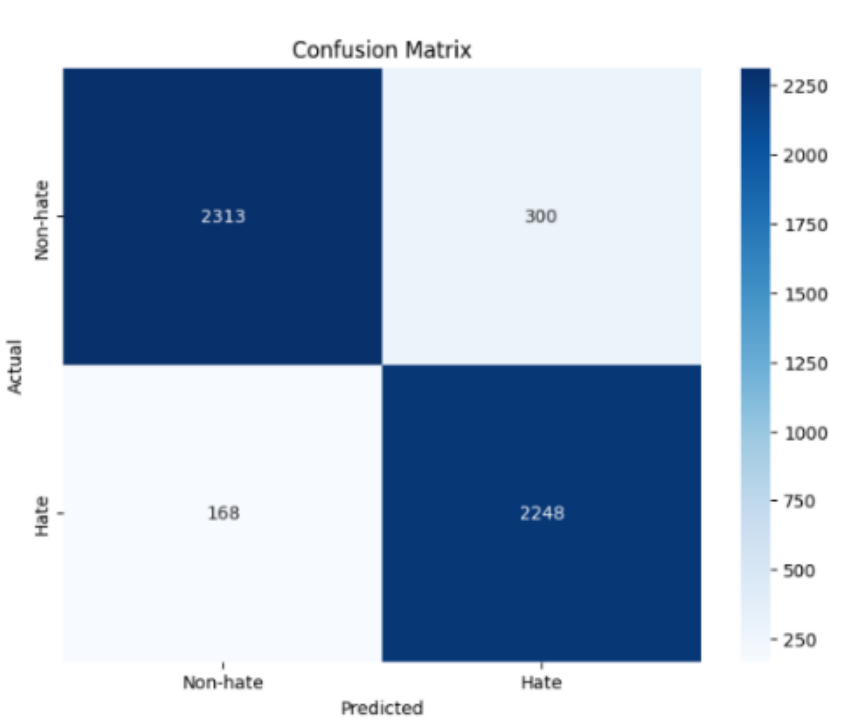

i) BanglaBERT + CNN on BDSHS Dataset

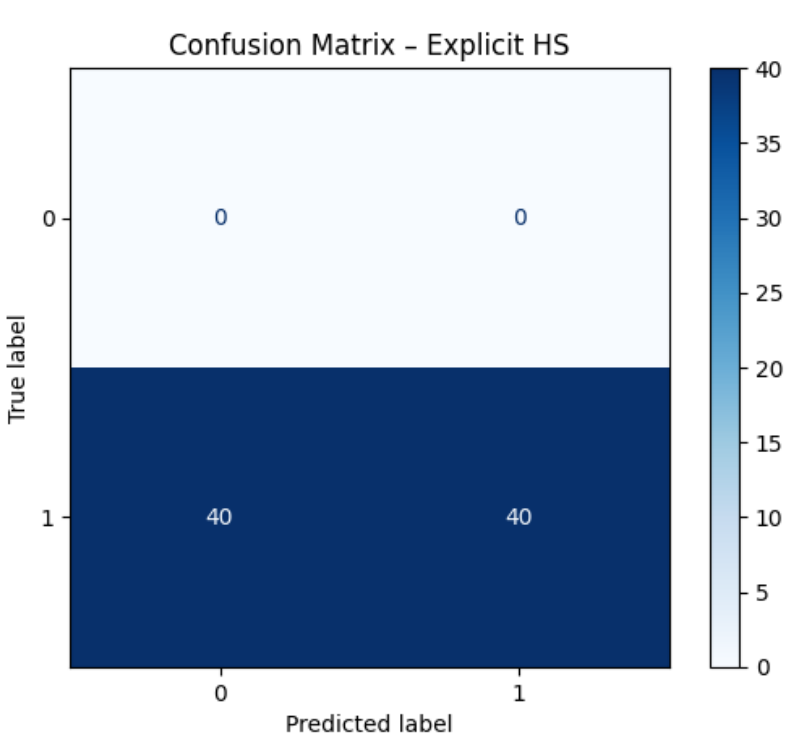

j) BanglaBERT + CNN (BDSHS) on Explicit HS

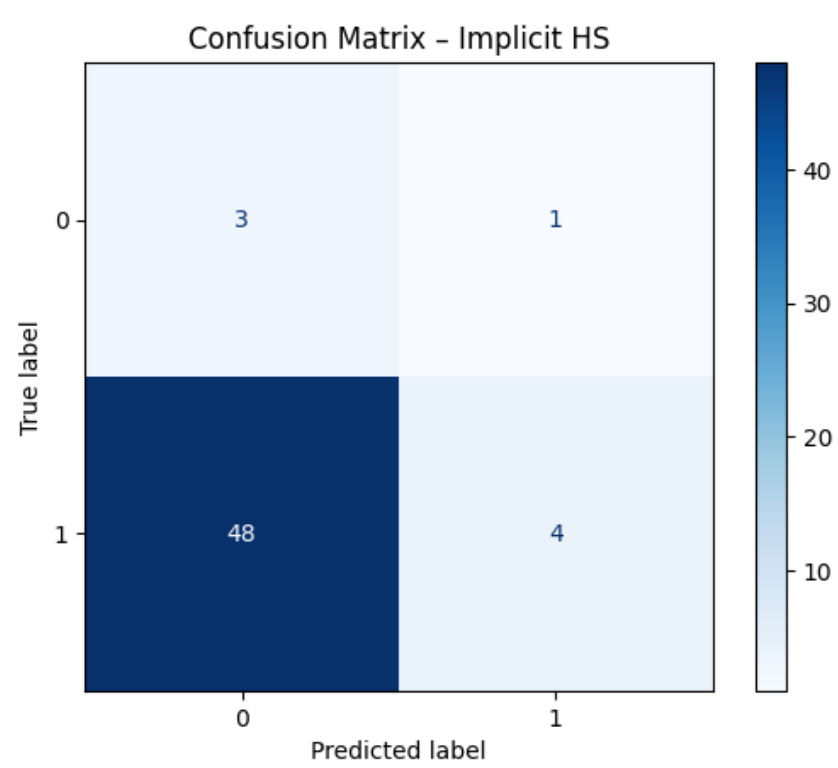

k) BanglaBERT + CNN (BDSHS) on Implicit HS

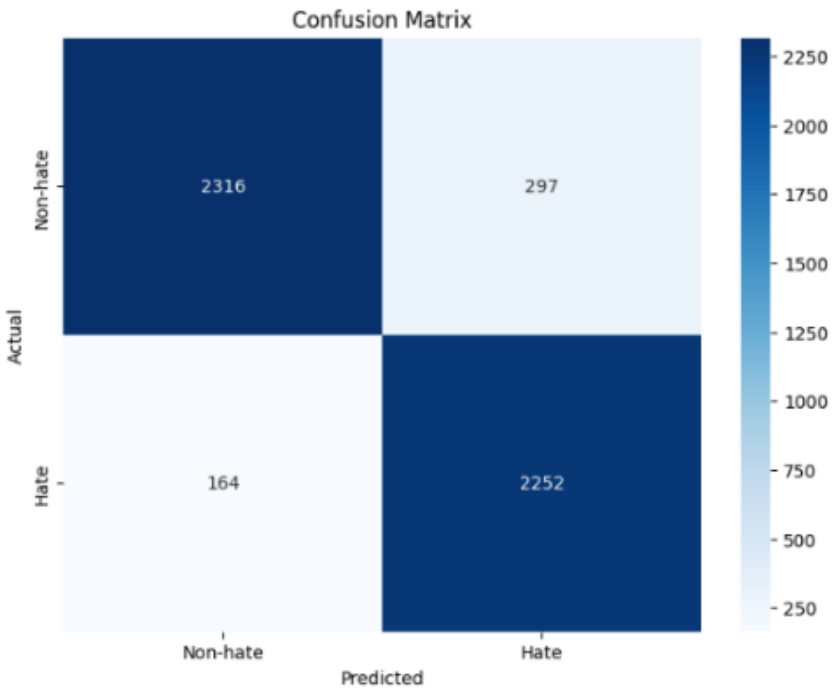

l) BanglaBERT + Bi-LSTM on BDSHS dataset

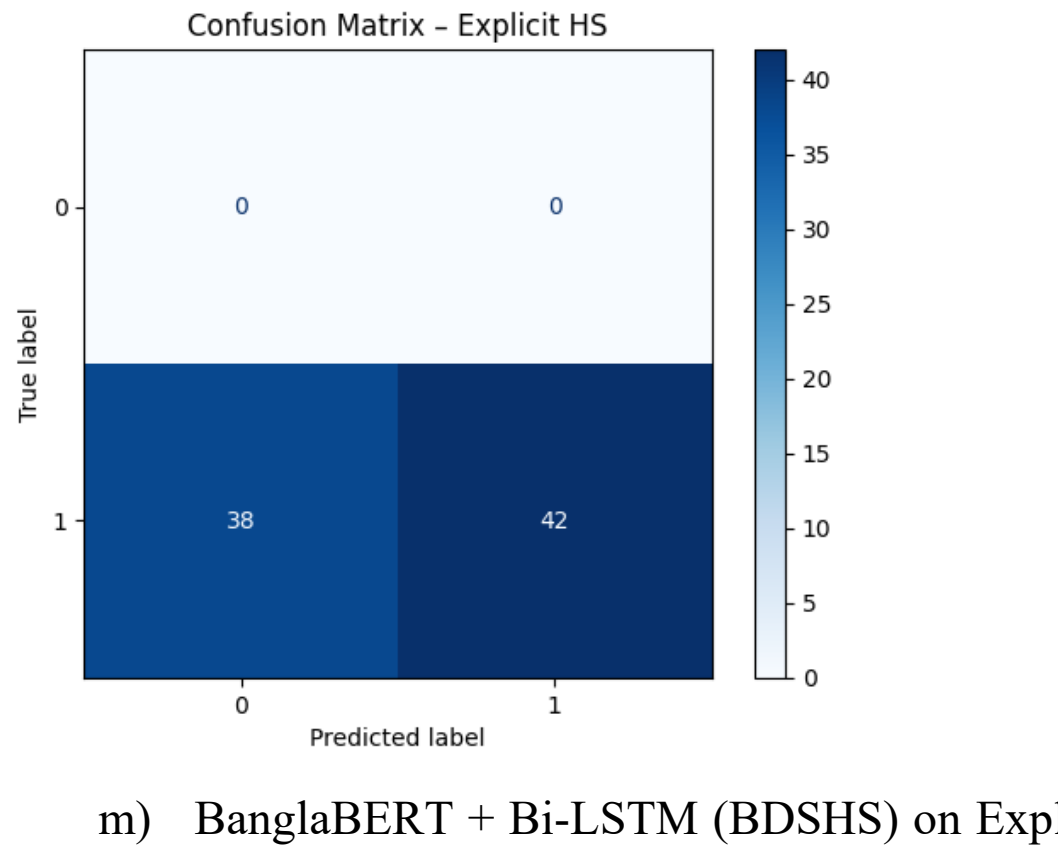

m) BanglaBERT + Bi-LSTM (BDSHS) on Explicit HS

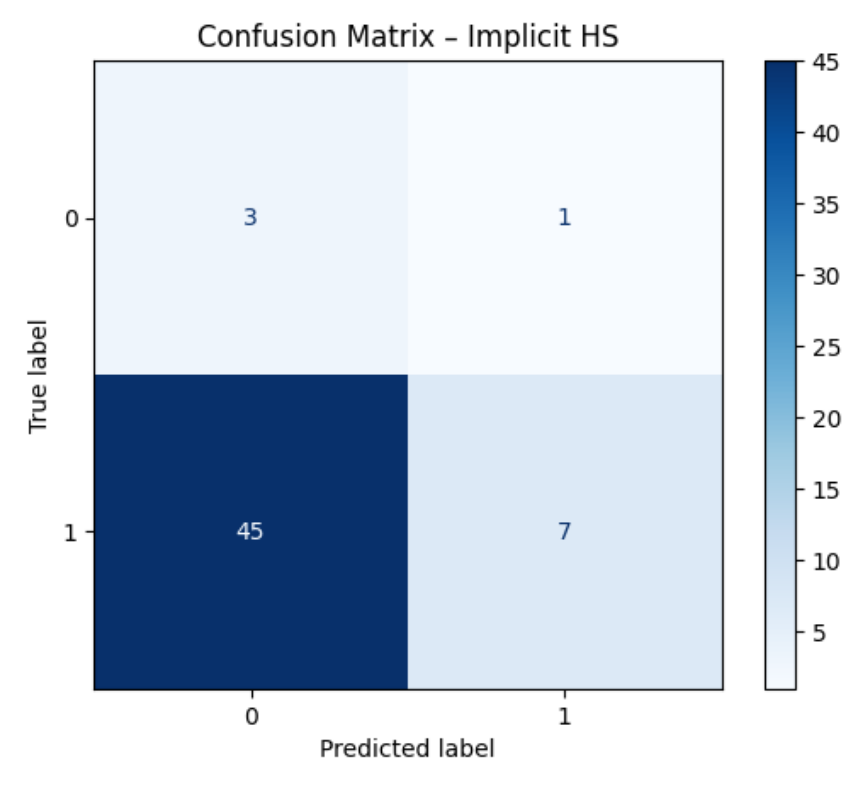

n) BanglaBERT + Bi-LSTM (BDSHS) on Implicit HS

**Fig. 3** Confusion matrices of BanglaBERT-based and hybrid models evaluated on benchmark, merged, and external datasets, including analyses on explicit and implicit HS.

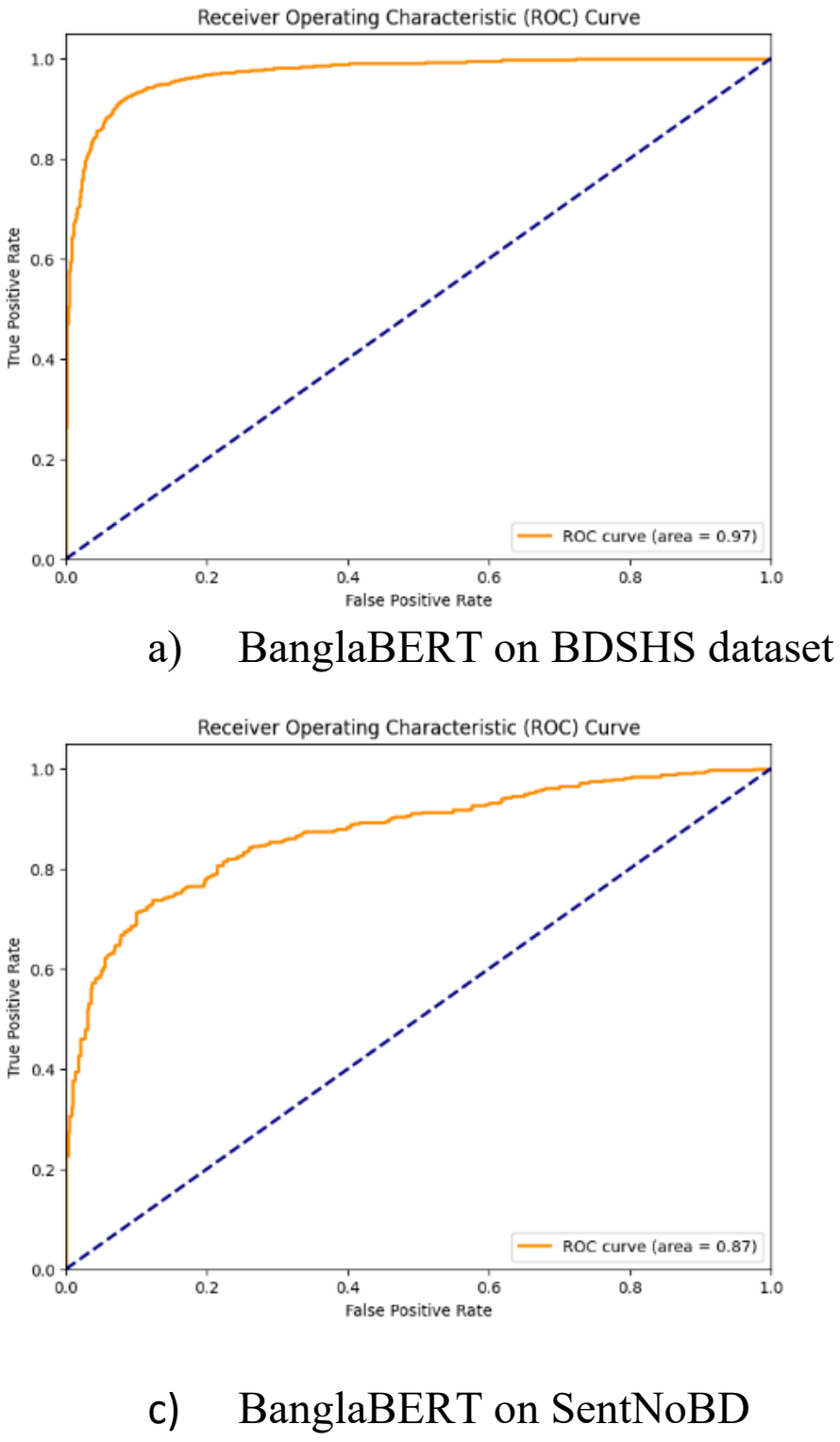

a) BanglaBERT on BDSHS dataset

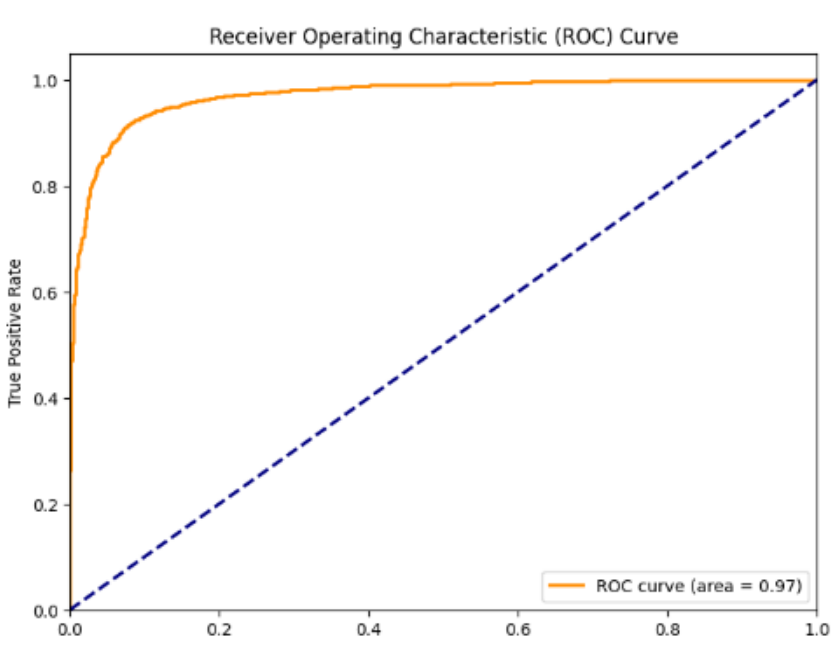

b) BanglaBERT on NNTI dataset

c) BanglaBERT on SentNoBD dataset

**Fig. 4** ROC curves of BanglaBERT-based model performances fine-tuned on 3 different benchmark datasets.

## *5.2 Generalization to Real-World (External Dataset)*

All the models performed significantly better on explicit HS, which typically involved clear instances of profanity, slurs, or aggression. However, their performance notably dropped when dealing with implicit HS ( → Fig. 5), where sarcasm, cultural innuendo, or indirect sentiment was used. Among the tested models,

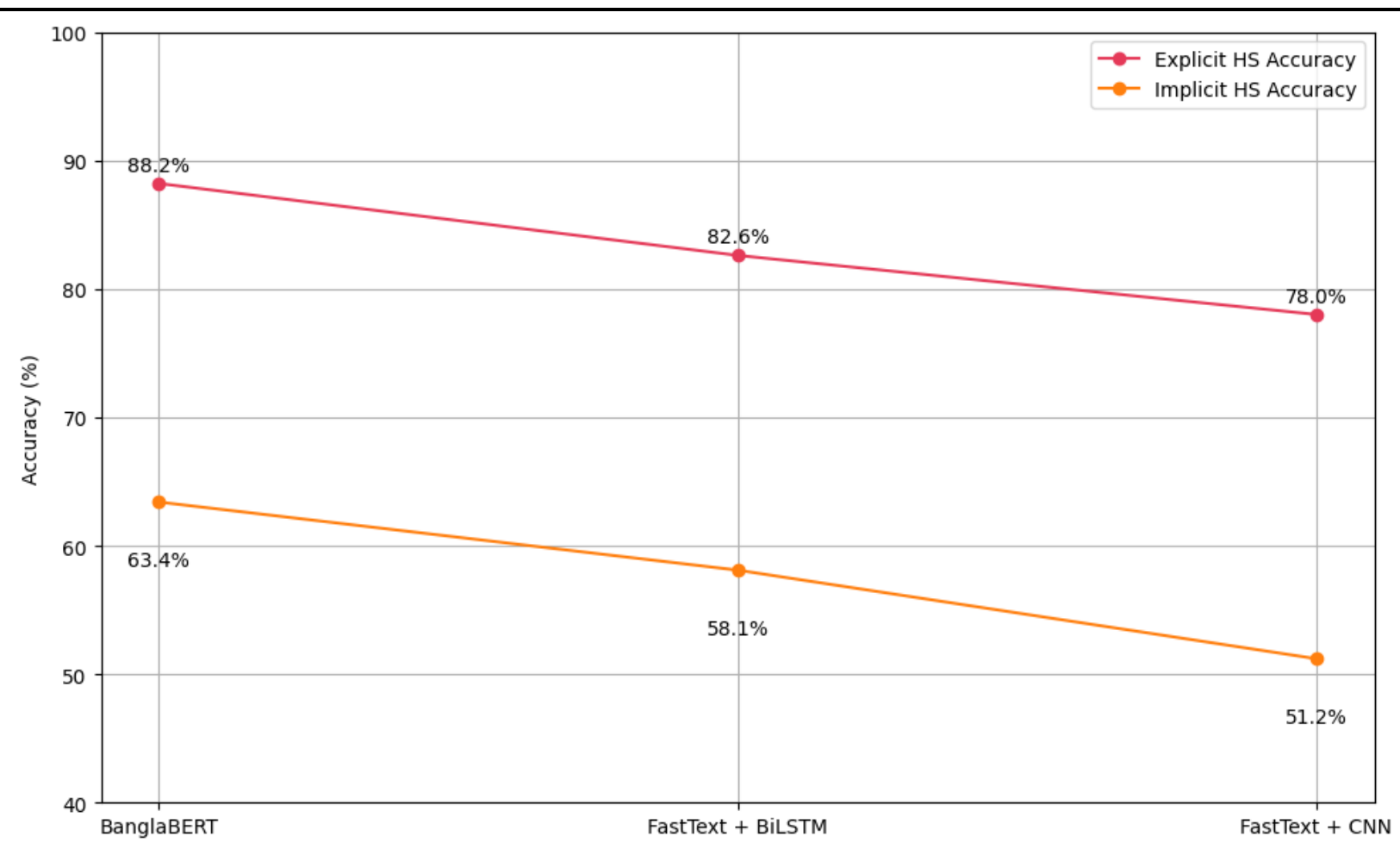

**Fig. 5** Comparison of model performance on explicit vs implicit HS detection

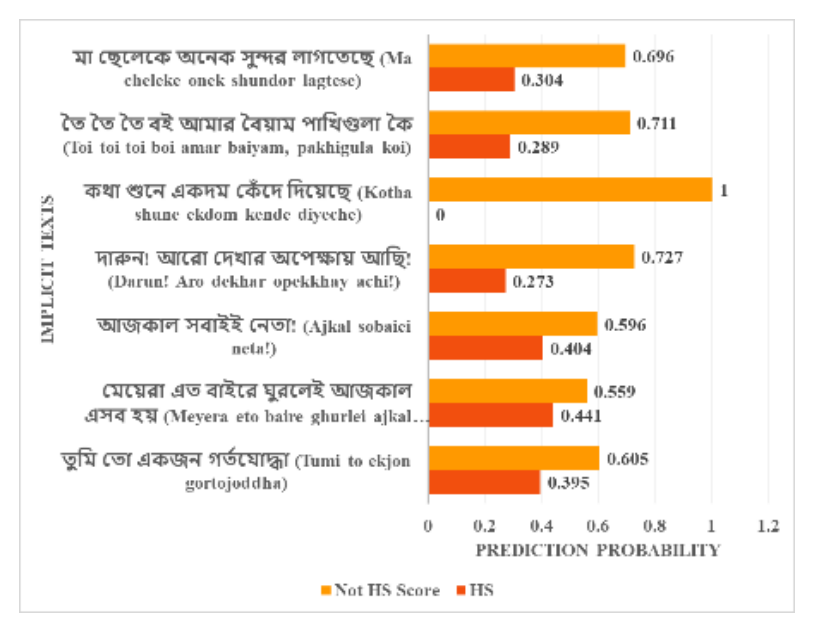

(a)

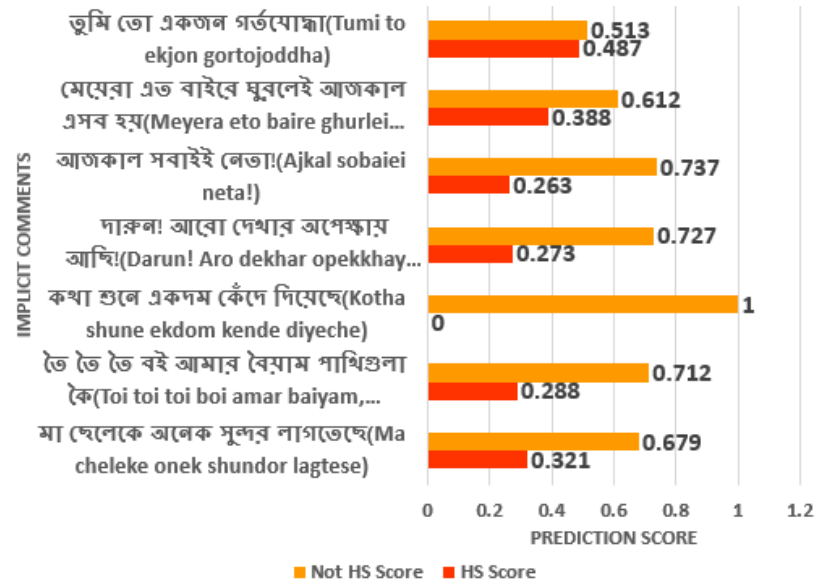

(b)

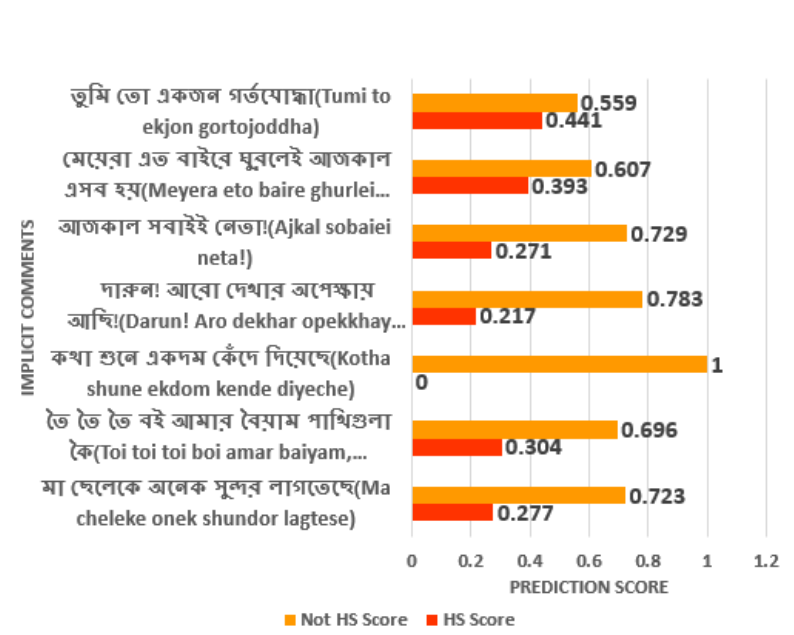

(c)

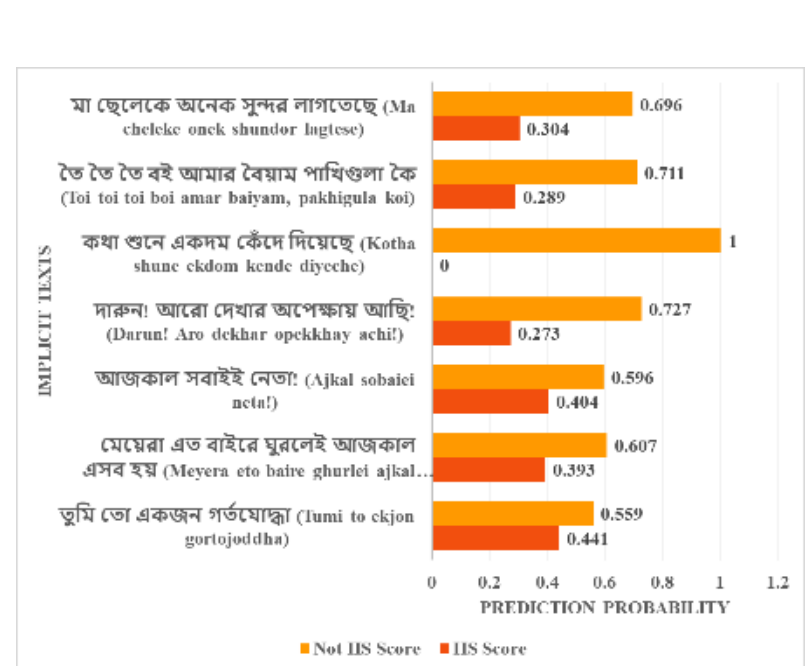

(d)

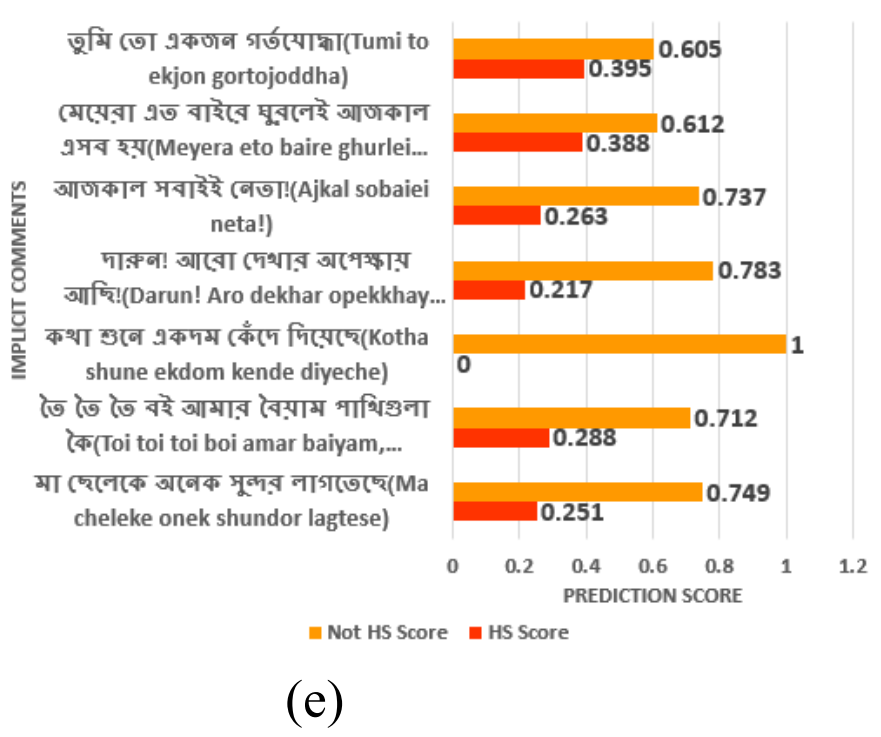

(e)

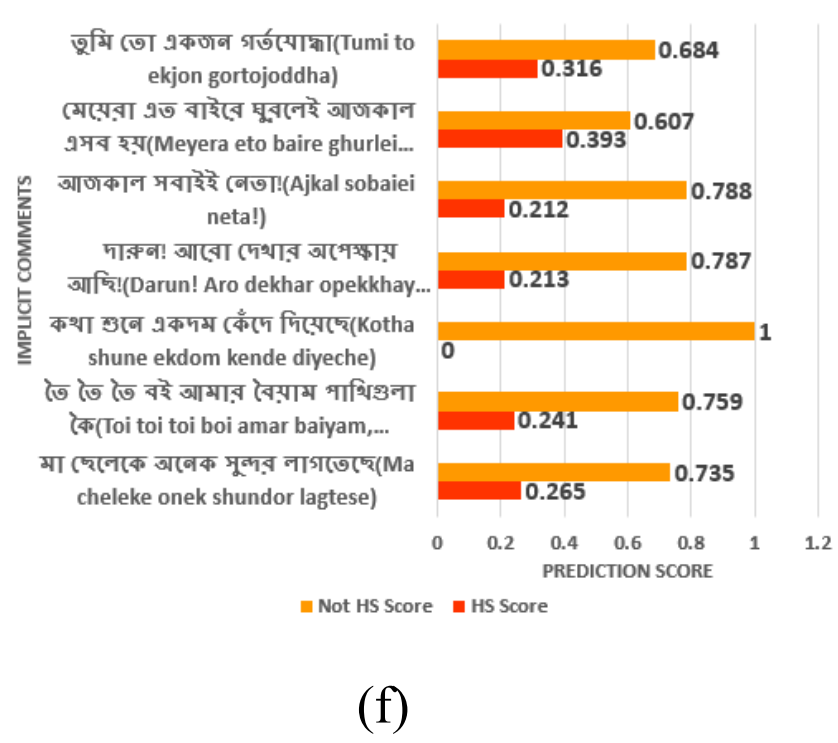

(f)

**Fig. 6** Model predictions on implicit HS examples. Each subplot shows prediction probabilities for each class across six architectures. The visualized examples include sarcasm, irony, satire, and culturally nuanced expressions that pose challenges for Bangla NLP systems. (a) BanglaBERT, (b) BanglaBERT+CNN, (c) BanglaBERT+BiLSTM, (d) FastText+BiLSTM, (e) FastText+CNN, and (f) Fast-Text+LSTM.

BanglaBERT demonstrated the best generalization capability—likely due to its contextual language understanding—although it still struggled to capture the subtleties of implicit expressions. All the models were tested on a manually labelled external dataset containing both explicit and implicit HS to evaluate their generalizability and robustness in real-world scenarios.

## *5.3 Qualitative Insights*

While quantitative metrics offer useful benchmarks, they often overlook the real-world complexities of HS detection—especially when speech is implicit, sarcastic, emoji-driven, or context-dependent. This section presents a qualitative analysis of the models, based on our curated external dataset. These aspects were explored in terms of (i) implicit HS, (ii) ambiguous interpretation of the same content (iii) emoji-based comments and their emotional implications, and (iv) the impact of text summarization on model predictions. These insights reveal how current models perform with culturally nuanced and dynamic content beyond their training data.

### 5.3.1 Implicit Hate Speech

Despite being dehumanizing, certain sentences lack overt profanity and are frequently misclassified (→ Fig. 6). This section examines failure cases across these models. Five recurring themes were identified where these models underperformed due to linguistic subtleties, lack of contextual grounding, and socio-cultural complexities. Each theme includes examples, illustrating the models' shortcomings in identifying implicit or culturally encoded HS.

**Table 13**: Examples of implicit HS in Bangla with Romanized and English translations.

| Category | Bangla (Romanized) | English Translation |
|---|---|---|
| Sarcastic / Indirect | তুমি তো একজন গর্তযোদ্ধা (Tumi to ekjon gortojoddha) | You're such a cave warrior |
| | আজকাল সবাই নেতা (Ajkal sobaiei neta) | Nowadays everyone is a leader |
| Victim-Blaming | মেয়েরা এত বাইরে ঘুরলেই (Meyera eto baire ghurlei) | Girls roaming = blame |
| Ambiguous Positivity | দারুন, আরো দেখার অপেক্ষা (Darun, aro dekhar opekkha) | Great, waiting to see more |
| | মা ছেলেকে অনেক সুন্দর লাগছে (Ma cheleke onek shundor lagtese) | Mother looks beautiful to son |

| Category | Bangla (Romanized) | English Translation |
|---|---|---|
| Meme / Trolling | তৈ তৈ তৈ বই আমার ব্যায়াম, পাখিগুলা কই (Toi toi toi boi amar baiyam, pakhigula koi) | Mocking meme |
| | কথা শুনে একদম কেঁন্দে দিয়েছে (Kotha shune ekdom kende diyeche) | "I cried a lot" meme |
| Ambiguous Trolling | এই তো আবার মুক্তিযুদ্ধের চেতনার ফেরিওয়ালা হাজির! (Ei to abar mukti juddher chetonar feriwala!) | Here comes another peddler of liberation war values! |
| | ওদের জন্য গুলি বরবাদ করিস না, একটা কুকুরই যথেষ্ট! (Oder jonno guli borbad koris na, ekta kukur e jothesto) | Don't waste bullets on them, one dog is enough! |

#### *5.3.1.1* Sarcasm

In Bangla, sarcasm often involves exaggerated praise or ridicule to convey contempt. An example is "Tumi to ekjon gortojoddha" (You're such a cave warrior), which, although contextually recognized as hateful, was assigned a low hate score by the models (BanglaBERT: 0.21; BiLSTM: 0.16; CNN: 0.18). Similarly, the phrase "Ajkal sobaiei neta!" (Nowadays everyone is a leader!') was also misclassified with low scores (BanglaBERT: 0.13; BiLSTM: 0.12; CNN: 0.10), highlighting the models' inability to detect historical and socio-political connotations in sarcasm → (Fig. 6a, b, c) .

#### *5.3.1.2* Meme-Based Expressions

Bangla meme culture often turns phrases that appear humorous into subtle insults. For instance, "Toi toi toi amar baiyam, pakhigula koi", once a phrase of empathy, is now a meme-based insult. The models misclassified it with low hate scores (BanglaBERT: 0.17; BiLSTM: 0.13; CNN: 0.11). Another example, "Kotha shune ekdom kende diyeche" (Cried right after hearing that), was similarly misclassified, revealing the models' inability to adapt to the evolving nature of meme language (→ Fig. 6a, b, c).

#### *5.3.1.3* Victim-Blaming

Certain expressions, while linguistically neutral, imply victim-blaming, particularly with gender biases. The phrase "Meyera eto baire ghurlei ajkal eshob hoy" ('This is what happens when girls roam outside too much') was not classified as HS by the models (BanglaBERT: 0.23; BiLSTM: 0.19; CNN: 0.14), (→ Fig. 6a, b, c) highlighting their failure to detect subtle, yet harmful, societal biases.

#### *5.3.1.4* Implicit Sexual Shaming

Some statements, although seemingly innocuous, carry sexual shaming undertones when placed in specific contexts. For example, "Ma cheleke onek shundor lagtese" (The mother looks very beautiful to her son) when posted under a woman's photo, takes on a mocking, sexual connotation. The models missed this context entirely, assigning low hate scores (BanglaBERT: 0.12; B-iLSTM: 0.08; CNN: 0.07), (→ Fig. 6a, b, c), indicating the importance of multimodal and contextual understanding.

#### *5.3.1.5* Ironic Praise

Some phrases appear complimentary but are used ironically to ridicule or mock, particularly in political discourse. An example is "Darun! Aro dekhar opekkhay achi!" ('Great! Waiting to see more!'), which the models mis-classified as non-hateful (BanglaBERT: 0.08; BiLSTM: 0.09; CNN: 0.07) (→ Fig. 6a, b, c). This demonstrates the models' reliance on sentiment polarity rather than the underlying irony or intent.

#### *5.3.1.6* Trolling

Bangla social media trolling mostly employs mockery or provocation, but does not often overtly hate words. For instance, the phrase "Ei to abar mukti juddher chet-onar feriwala" (Here comes another peddler of liberation war values!) was identified as HS by BanglaBERT (score: 0.76), although it is a sarcastic political comment rather than a hateful comment. Conversely, the sentence "Oder jonno guli nosto korish na, ekta kukur e jothesto" (Don't waste bullets on them, one dog is enough!) had a low hate score from FastText+BiLSTM (0.22), although it dehumanized a group (→ Fig. 6d, e, f). These instances show how models confuse trolling with hate or are unable to detect hate embedded in sarcasm.

### 5.3.2 Context-dependency: Ambiguous interpretation

In addition to trolling and sarcasm, many Bangla social media posts return ambiguous sentiment, wherein a sentence can be interpreted as genuine praise or sarcastic mockery based on context. Examples also demonstrate how even surface-level sentiment fails to represent actual intent. → Table 12 provides such context-dependent postings with their opposite interpretations. Our research revealed the greatest obstacles to Bangla HS detection: semantic shifts in trending terms, ambiguous language masking hostility, lack of contextual cues such as tone or source posts, and static models that fail to keep up with evolving online language. In each category, the models' inability to interpret contextual and cultural nuances led to the misclassification of sarcastic remarks, meme-based ridicule, and socially normalized biases as neutral. These shortcomings highlight the need for models to incorporate culturally contextualized datasets, adapt dynamically to slang, and employ multimodal modeling to enhance the detection of implicit HS.

**Table 14:** Context-dependent Bangla comments with ambiguous interpretations

| Bangla Comment | Possible Interpretations |
|---|---|
| তোমার মতো হলে দেশটা বদলাতো! (Tomar moto hole deshta bodlato!) | Could express genuine praise on a post about social service, or sarcasm on a post by a controversial figure. |
| অনেক কিছুই শিখলাম আজ! (Onek kichui shikhlam aaj!) | Could be genuine appreciation on an educational post, or mockery on misleading or self-important content. |
| তোমার আত্মবিশ্বাস দেখেই তো দেশ আজ এই জায়গায়! (Tomar atmobishwas dekhei to desh aj ei jaygay!) | Could sound motivational on a patriot's post, or bitterly sarcastic on a politician's post blamed for failure. |
| এত বুদ্ধি থাকলে নোবেল পেতে পারতেন! (Eto buddhi thakle Nobel pete parten!) | May be literal praise on insightful content, or sarcasm aimed at a poor argument. |
| এই কথা শুধু আপনি-ই বলতে পারেন! (Ei kotha shudhu apni-i bolte paren!) | Can signal respect if said to a wise speaker, or sarcasm if said to a controversial figure. |
| অভিনন্দন, সাহসটা শুধু আপনার আছে! (Abhinondon, sahosta shudhu apnar achhe!) | Could be applause for bravery, or mockery if the act was reckless. |
| এই পোস্টে বস অনেক কিছু বলেছে, বুঝলে বুঝলে! (Ei poste boss onek kichu boleche, bujhle bujhle!) | Could show support for bold speech, or act as a coded insult referring to passive aggression. |

### 5.3.3 Emoji-based Comments and Sentiment Masking

Emojis carry emotional cues that modify or intensify a comment's tone. In this analysis, emojis were found to be used frequently to express sarcasm, mockery, or hidden aggression. Even factual sentences could be employed to hate if paired with the right emoji. For instance, "তুই কিছুই পারিস না" 😡🔥expresses aggression, whereas the same sentence with 😂appears to be teasing (→ Table 13). The models varied in the manner in which they handled such signals. When emojis were retained and translated into sentiment tokens (e.g., 🤮→"ঘৃণাজনক"(disgusting), BanglaBERT would at times capture the implied hate. However, lighter models such as FastText+Bi-LSTM and CNNs would typically ignore or drop emojis as part of preprocessing, leading to misclassification. Furthermore, a single emoji can convey hugely disparate meanings depending on context—humor, threat, or admiration—so that static translations are unreliable. This kind of ambiguity is seen in → Fig. 7, where emojis flip model predictions. Models based on BanglaBERT perform better than models based on FastText in identifying Bangla HS, especially in the case of implicit and context-dependent HS ( → Fig. 7a). BanglaBERT + BiLSTM achieved the highest accuracy (65% TP) when both sequential and contextual modelling were used. BanglaBERT + CNN was a close second, and the base BanglaBERT also worked effectively. For the FastText models, the LSTM model worked slightly better than the BiLSTM and CNN, although all the models fared poorly in terns of fine-grained hate. Surprisingly, FastText + CNN issued the fewest false alarms but the poorest true positive rate. In general, the combination of BanglaBERT with BiLSTM offers the best well-balanced detection. To quantify emoji impact, model performance was compared on two test sets: one with emojis translated with bnemo, and one with emojis removed ( → Fig. 7b). Emoji removal consistently reduced F1-scores, with BanglaBERT being affected the most. This confirms that emojis, although informal, provide a critical emotional context that models leverage to detect implicit hate. Nonetheless, without contextual comprehension, models fail, highlighting the need for more fine-grained emoji-aware processing in Bangla NLP. Hybrid models such as BanglaBERT + BiLSTM benefit the most from the emoji context, considering their decent sequential and contextual understanding, with the greatest F1-score boost. BanglaBERT + CNN also benefits, but not as much so. FastText-based models benefit less overall, as they do not use emoji cues effectively, especially when not pre-trained. This supports the fact that, more deeply, contextually aware architectures utilize emojis more beneficially for detecting implicit HS.

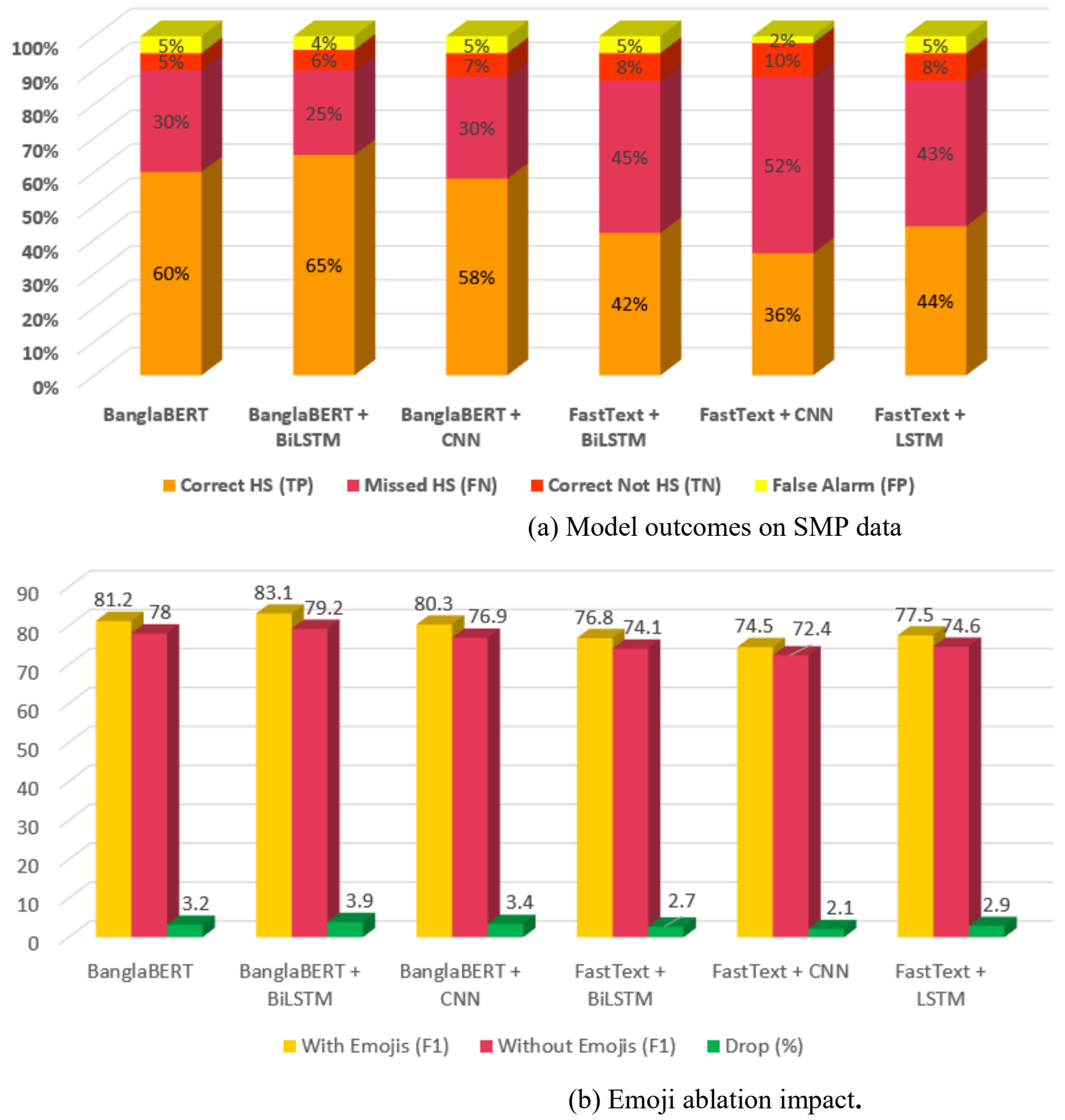

(a) Model outcomes on SMP data

(b) Emoji ablation impact.

**Fig. 7** Impact of emojis on Bangla HS detection: (a) model prediction breakdown on real-world comments, and (b) performance shifts after removing emoji content.

### 5.3.4 Text Summarization

A straightforward summarization technique, i.e., TextRank (Mihalcea & Tarau, 2004), was used to reduce the input size of lengthy comments, which led to a minor performance improvement of 1–2% in the F1 score by effectively eliminating irrelevant or noisy content. However, this came at the cost of sacrificing some context information that was necessary in conveying HS, particularly where more sophisticated or covert information was lost in a summary.

### 5.3.5 Over-policing vs. Free Speech

While the models compared in this study demonstrated promising accuracy in detecting both explicit and implicit HS, our qualitative analysis revealed an important ethical concern: the risk of over-policing speech, particularly in politically sensitive or satirical contexts. For instance, benign comments using words like ``আপা" (a respectful term for a female leader) or ``আপারভাই" may be misclassified as HS due to the presence of political references, even when the actual intent is criticism or satire. Such misclassifications can suppress legitimate political expression and infringe on users' freedom of speech. This challenge is (→ Table 13) especially critical in low-resource languages such as Bangla, where socio-political nuances, dialectal variation, and cultural idioms complicate HS detection. An overreliance on surface-level patterns and insufficient contextual understanding in current models may lead to false positives, ultimately eroding trust in moderation systems.

This study recommends that future work incorporate human-in-the-loop systems, context-aware annotation practices, and an emphasis on distinguishing satire and criticism from actual HS. This balance is crucial to ensuring that HS detection systems support safer discourse without stifling democratic participation.

**Table 15**. Examples of implicit HS with an emoji context and model interpretation.

| Bangla/Romanized | English Translation | Emoji | Intended Tone | Model Prediction | Actual Meaning |
|---|---|---|---|---|---|
| তুই কিছুই পারিস না (Tui kichui paris na) | You cannot do anything | 😡 🔥 | Aggressive insult | Hate | Direct HS with emotional aggression |
| তুই কিছুই পারিস না (Tui kichui paris na) | You cannot do anything | 😂 👏 | Teasing | Not HS | Sarcastic mockery; emotionally less intense |
| তোর মত মানুষদের জন্যই দেশটা শেষ (Tor moto manusher jonnoi deshta shesh) | Because of people like you, the country is doomed | 😠 🇧🇩 | Political hate | Hate | Strong accusatory tone; nationalist hate |
| তোর মত মানুষদের জন্যই দেশটা শেষ (Tor moto manusher jonnoi deshta shesh) | Because of people like you, the country is doomed | 😂 🤣 | Sarcastic trolling | Not HS | Mocking tone; hate masked as humor |
| এরা সব জাতির গর্ব (Era shob jatir gorbo) | These people are the pride of the nation | 🤢 😫 | Sarcasm / Disgust | Not HS | Disguised insult using sarcasm and disgust |
| সাবাস! এবার তোমাদের কবর দিতে হবে (Shabash! Ebar toder kobor ditei hobe) | Well done! Now you all must be buried | 🙂 | Threat (masked as humor) | Not HS | Hidden threat disguised as light humor |

## *5.4 Discussion*

### 5.4.1 Strength

This paper provides a detailed real-world analysis of current HS detection models in Bangla across multiple dimensions, such as benchmarking, data variation, implicit speech, emoji-driven sentiment, and cultural ambiguity. The model architectures themselves remain unchanged; however, no prior work has systematically and in a layered manner evaluated both the generalization and pragmatic failures of these models when applied to informal, noisy social media data, despite being trained on clean data. One of its main assets is to have used and developed a manually curated real-world test set enabled by implicit HS, emoji, or ways to sarcastically or codify the language. This approach allows for a realistic evaluation that goes beyond clean benchmark conditions and highlights the real challenges faced by current systems when dealing with context-variant Bangla expressions dealing with context-variant Bangla expressions. The evaluation findings indicated that BanglaBERT and similar models are highly effective on benchmark data but perform poorly on real-world data, particularly for implicit HS. This lack of uniformity underscores a deeper matter that the deep learning models generally used so much today are built to leap out from things such as shallow HSs and do not fare well with rather subtle forms of hate hidden under sarcasm, irony, memes, cultural references, and connotations. The presented qualitative analysis identified common mistakes in sarcastic compliments, victim-blaming, and ambiguous sentiment, emphasizing that there are deeper contextual representations. Another specific examination of emoji interpretation is a significant contribution. Emojis, which are widely employed by Bangladeshi netizens, frequently perform inverse or hyper amplification of a sentence's meaning, serving as indicators of sarcasm or intensifiers of emotion. Models that simply remove emojis tend to misclassify these posts. Even with its contextual superiority, BanglaBERT was also emoji-ambiguous in the absence of additional context. The present study demonstrated that the same Bangla sentence can be considered praise or insult by emoji use ( → Table 12), which indicates the inefficiency of considering only static text processing techniques. These observations indicate that lexical or keyword-based models would not be enough for analysing/identifying Bangla HS. There are several actionable directions forward, as suggested by the results: (i) emoji-aware preprocessing, (ii) training on sarcastic and sentiment-ambiguous data, and (iii) incorporation of discourse- or multimodal-level features. By using an annotated, externally generated real-world test suite, we are able to effectively demonstrate that even the strongest models can be fooled, and intuitive knowledge concerning cultural and social context is not only beneficial but indeed mandatory for Bangla HS detection algorithms to be trusted. For models to be trusted in such real-time moderation scenarios, they need to be sensitive to linguistic nuance and the evolving nature of discussions on the internet.

### 5.4.2 Limitations

Despite the novelty and practical relevance of our study on Bangla HS detection, there are some limitations in this study which should be addressed in future work. The external dataset, while manually collected to contain implicit hate and emoji-rich expressions, is small (~200 samples). This limits the capacity to perform truly expressive, robust category-wise analysis (e.g., across political vs. religious hate) or to train models end-to-end using entirely informal examples. The results are mainly diagnostic and comparative, rather than generative or generalizable to low-resource settings more widely. Although the emoji assignation framework is interesting, it relied on static translation methods rather than bnemo methods. The experiment demonstrated that emoji-aware preprocessing enhances performance, but failed to create dynamic, context-dependent emoji interpretability, which could be essential for eliminating sentiment ambiguity. The results demonstrate that the extant models do not perform well under these conditions, but this leaves new multimodal, contextual emoji processing techniques for others to investigate. Finally, while we carefully discussed misclassification, over-policing, and misinterpretation of culture in this paper, this assessment was text-based. More sophisticated multi-lexical extensions, including the use of images, conversation history, or user metadata, can be used to help disambiguate sarcasm, satire, and coded language, but these multimodal extensions can be saved for future work. For realistic benchmarking, baselines based on existing hybrid architectures (BanglaBERT + CNN, BiLSTM) were used as this research did not introduce or tune new architectures. Therefore, these findings are limited to investigating of how well current architectures can generalize to real-life Bangla SMP data and how future architectures could be improved to better fit and solve the task. The current approach determines only Bangla characters and the embedded hate vernaculars, but it can also be similarly utilized for different languages. To generalize this methodology to other languages, it is necessary to consider both the architectural and dataset levels. From a modeling perspective, this setup would need to extend from monolingual embeddings (e.g., BanglaBERT) to multilingual setups such as mBERT, XLMR, or language-specific fine-tuned transformers to encapsulate cross-lingual syntactic and semantic differences. At the dataset level, the strategy depends on whether the joined languages are high-resource or low-resource languages. For high-resource languages such as English or Arabic, large HS-annotated corpora are already available, which would only require adapting the preprocessing pipeline (emoji handling, sarcasm-aware tokenization, and code-mixing treatment) to the new language and fine-tuning the model on those corpora. For low-resource languages, that challenge is even more foundational. New datasets need to be created by (i) having native speakers participate in crowdsourced annotation efforts, (ii) bootstrapping from high-resource languages through cross-lingual transfer using multilingual embeddings, (iii) semi-automated corpus construction—e.g., via translation or synthetic augmentation with larger language models—and (iv) culture-sensitive labeling guidelines that address local sarcasm, memes, and coded hate expressions. These are necessary in order to guarantee linguistic coverage as well as cultural validity. Accordingly, Bangla is considered here as an implementation case of low-resource, and the higher-level methodology will be able to be extended to a mixed high- and low-resource multilingual HS detection in future work.

### 5.4.3 Performances Comparison with related recent work

Compared with recent works (e.g., BanglaHateBERT and DeepHateExplainer), the present research provides a broader and deeper evaluation. Although previous methods have performed well on benchmark datasets (with F1-scores in the range of 0.72–0.86), few have explored the performance of such systems in the wild, such as those subjected to emoji-driven sentiment masking or sarcastic trolling. In contrast, the present study shows that on the benchmark data, even the basic vanilla BanglaBERT outperforms the top-performing model by 91.4% F1-score, whereas when evaluated on noisy, real-world, emoji-rich samples, it scores 75.3% F1—a severe puzzlement, to say the least. The segmented understanding of BanglaBERT's performance identifies strong and weak aspects. Its contextualized representations are very effective at detecting explicit hate, profanities, and direct abuse. Nonetheless, the transformer's attention mechanism alone fails to infer the pragmatic cues hidden in informative mimicry, ironic compliments, or culturally based slang. Models such as BanglaBERT + BiLSTM are only able to partially fill this gap by capturing the sequential relationship and, consequently, have a better tolerance against sentiment ambiguity and indirect expression. Note that even the best-performing systems do not reach 65% per HS across all sub-tasks, showing that there is still room for improvement in multimodal signalling, context chaining, and discourse-aware modelling. These observations pave the way for future directions of integrating information from external knowledge graphs, such as knowledge bases, conversation history features, or user-level metadata, to better understand sarcastic or coded hate for detecting HS in a Bangla setting.

**Table** 16: Examples of Politically Sensitive Non-HS Comments Misclassified as Hate

| Bangla Texts (Romanized) | Translation | Predicted Label | Human Annotation |
|---|---|---|---|
| আপার ভাইরে কেমন আছেন? (Apar bhaire kemon achen) | How is sister's brother doing? | HS | Not HS (Political Criticism) |
| তোমার আত্মবিশ্বাস দেখেই তো দেশ আজ এই জায়গায় (Tomar atmobishash dekhei to desh aaj ei jaigay) | Just by seeing your confidence, the country has reached this point today. | HS | Not HS (Sarcastic Commentary) |
| মামুনের মা কি বলে (Mamuner ma ki bole) | What does Mamun's mother say? | HS | Not HS (Meme/Humor) |

These examples highlight the tendency of existing models to over-police political and cultural satire

## *5.5 Theoretical and Practical Implications*

A core theoretical contribution of this study lies in demonstrating that models like BanglaBERT, despite performing well on benchmark datasets, fail to generalize to real-world data that contains implicit hate speech, sarcasm, and emojis. This underscores a significant gap in the existing literature, which often focuses on models' performance in controlled environments but does not account for the complexities of real-world data. Our findings suggest that deep learning models need to account for contextual language and non-explicit expressions to truly perform well in live, informal social media environments. Implicit hate speech, which can be conveyed through sarcasm, irony, and cultural references, remains a challenging area for automated systems. Our study highlights that current SOTA models, while proficient in detecting explicit hate speech, struggle with nuanced hate and often fail to interpret context-dependent cues. This provides a new theoretical understanding of the limitations of existing models and calls for further exploration into multimodal, contextual, and emotion-aware processing for future models. The real-world implications for social media moderation systems. The results emphasize the need for models that are sensitive to linguistic nuance and capable of detecting implicit hate speech in dynamic online conversations. Emoji-aware preprocessing, sarcasm detection, and contextual understanding must be incorporated into hate speech detection systems to ensure they can handle complex online content effectively.

## 6 Conclusion & Future work

In this research, a comprehensive comparative assessment of recent Bangla HS classification models was performed on three types of datasets: benchmark corpora, an integrated dataset, and a real-life instance from Bangladeshi social media. By highlighting the shortcomings of SOTA models on real-world data, particularly in the detection of implicit HS, this work encourages the development of more context-sensitive and culturally aware systems for online content moderation. While models such as BanglaBERT performed extremely well on sanitized datasets, their performance significantly decreased on informal, implicitly abusive, and emoji-rich text. Models such as BanglaBERT achieved strong performance on sanitized benchmark datasets, with accuracies reaching 91.4% and macro-averaged F1-scores as high as 0.91. However, when they are evaluated on informal, implicitly abusive, and emoji-rich texts, their performance dropped significantly. For example, BanglaBERT's combined accuracy decreased by approximately 16% (from 91.4% to 75.3%), and the F1-scores decreased similarly, highlighting the challenge of handling noisy, context-dependent online language. These findings reflected the limitations of current methods when they are applied to noisy, context-dependent online text. Qualitative research also revealed excellent blind spots in model reasoning, particularly with sarcasm, ironic complimenting, and emoji commentary, where hate prediction scores typically fell below 0.20 and true positive rates fell more than 30% compared with explicit cases. These findings underscore the importance of incorporating cultural, emotional, and contextual knowledge into Bangla HS detection systems. Future work should include designing guidelines for annotations that better identify implicit and indirect hate, real-world datasets for capturing the dynamic aspects of social media language, and emoji-aware preprocessing techniques. Multimodal approaches that combine text with images or metadata and explainable AI approaches also have the potential to make models more interpretable and trustworthy. Finally, the deployment of lean, real-time systems that respond to the dynamic aspects of online language is a necessary goal. It is anticipated that this work will spur more effort toward ethically grounded, culturally sensitive HS detection for low-resource languages such as Bangla.

**Author Contribution statements:** **Faria Afrin Tisha:** Writing – original draft, Writing – review & editing, Visualization, Methodology, Investigation, Data curation, Conceptualization; **Fariya Tabassum:** Writing – original draft, Visualization, Methodology, Investigation, Data curation, Conceptualization; **Hafsa Binte Kibria:** Writing – review & editing, Methodology, Investigation, Visualization, Validation, Supervision, Data curation; **Md Nahiduzzaman:** Writing – review & editing, Validation, Supervision, Formal analysis. **Mominul Ahsan:** Writing – review & editing, Visualization, Validation, Supervision, Methodology, Formal analysis.

**Data Availability Statement:** The data presented in this study are available in the article.

**Acknowledgments:** Throughout the writing process, the authors utilized AI technologies (e.g., ChatGPT) to refine the document's language and enhance its readability. After applying these technologies, they carefully assessed and revised the text as needed. The authors are solely responsible for the fundamental research, findings, and outcomes presented in this paper.

**Declaration of Competing Interest:** The authors affirm that they do not possess any identifiable conflicting financial interests or personal relationships that might have been perceived to impact on the findings presented in this paper.